\documentclass[10pt,twocolumn,letterpaper]{article}

\usepackage{cvpr}              % To produce the CAMERA-READY version
\usepackage{listings}
\usepackage{multirow}
\usepackage{colortbl}
\usepackage{algorithm}
\usepackage{algorithmic}

\definecolor{cvprblue}{rgb}{0.21,0.49,0.74}
\usepackage[pagebackref,breaklinks,colorlinks,allcolors=cvprblue]{hyperref}
\usepackage{graphicx}
\title{Enhanced OoD Detection through Cross-Modal Alignment of Multi-Modal Representations}

\author{Jeonghyeon Kim \: Sangheum Hwang\thanks{Corresponding Author}\\
Seoul National University of Science and Technology\\
{\tt\small \{mawjdgus, shwang\}@
seoultech.ac.kr}
}

\begin{document}
\maketitle
\begin{abstract}
% original version
Prior research on out-of-distribution detection (OoDD) has primarily focused on single-modality models. Recently, with the advent of large-scale pretrained vision-language models such as CLIP, OoDD methods utilizing such multi-modal representations through zero-shot and prompt learning strategies have emerged. However, these methods typically involve either freezing the pretrained weights or only partially tuning them, which can be suboptimal for downstream datasets. In this paper, we highlight that multi-modal fine-tuning (MMFT) can achieve notable OoDD performance. Despite some recent works demonstrating the impact of fine-tuning methods for OoDD, there remains significant potential for performance improvement. We investigate the limitation of na\"ive fine-tuning methods, examining why they fail to fully leverage the pretrained knowledge. Our empirical analysis suggests that this issue could stem from the modality gap within in-distribution (ID) embeddings. To address this, we propose a training objective that enhances cross-modal alignment by regularizing the distances between image and text embeddings of ID data. This adjustment helps in better utilizing pretrained textual information by aligning similar semantics from different modalities (i.e., text and image) more closely in the hyperspherical representation space. We theoretically demonstrate that the proposed regularization corresponds to the maximum likelihood estimation of an energy-based model on a hypersphere. Utilizing ImageNet-1k OoD benchmark datasets, we show that our method, combined with post-hoc OoDD approaches leveraging pretrained knowledge (e.g., NegLabel), significantly outperforms existing methods, achieving state-of-the-art OoDD performance and leading ID accuracy.

\end{abstract}

\vspace{-10pt}
\section{Introduction}
\label{sec:intro}

In open-world deployments, detecting out-of-distribution (OoD) data is crucial for ensuring the reliability and safety of machine learning models. Without robust out-of-distribution detection (OoDD), models may produce erroneous decisions that could lead to severe consequences in various applications such as medical diagnosis and autonomous driving.
Therefore, it is imperative that models not only accurately classify in-distribution (ID) data but also effectively identify and respond to OoD inputs.

%-------------------------------------------------------------------------

\begin{figure*}[htp]
    \centering
    \begin{subfigure}{.49\linewidth}
        \centering
        \includegraphics[width=\linewidth]{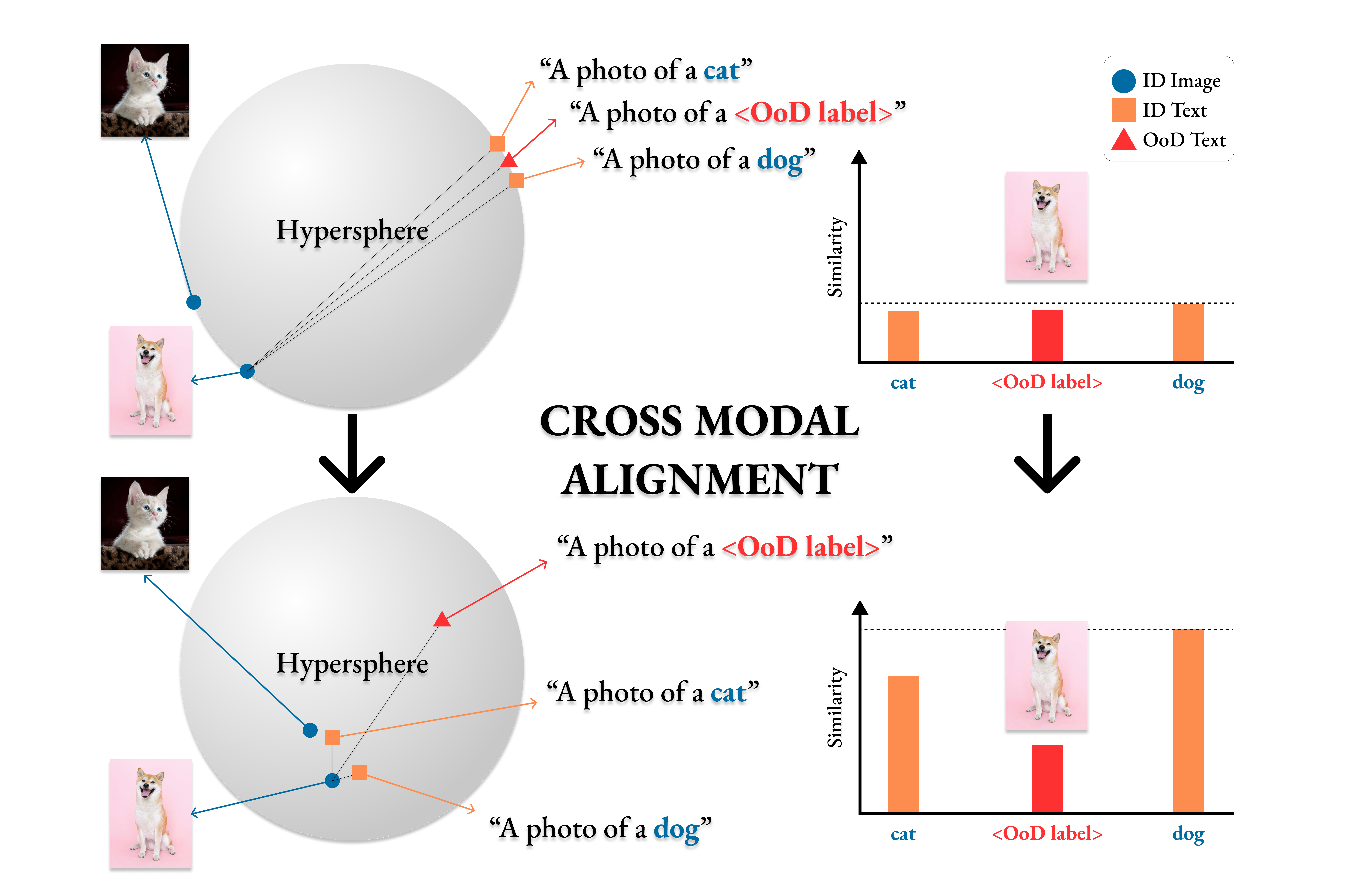}
        \caption{Illustration of CMA}
        \label{Fig1: a}
    \end{subfigure}
    \begin{subfigure}{.49\linewidth}
        \centering
        \includegraphics[width=\linewidth]{assets/main-b.pdf}
        \caption{MOS benchmark: ID ACC vs. 1-FPR95}
        \label{Fig1: b}
    \end{subfigure}
    \caption{(a) illustrates the hyperspherical embedding space and the corresponding cosine similarity values between the ``dog'' image and ``A photo of a $<\textit{label}>$'' text embeddings. Initially, the embedding space shows a bipartite separation between images and texts (top)~\cite{liang2022mind,oh2024geodesic, shi2023towards}. Through CMA, ID images and texts are brought closer together while maintaining a clear separation from OoD texts (bottom). This alignment enhances the discriminability of ID data from negative concepts (i.e., OoD labels), thereby improving OoDD performance. In (b), uncolored shapes represent MCM, while colored shapes denote NegLabel. The arrows indicate the effect of NegLabel compared to MCM, demonstrating that our method enhances its effectiveness.
    Points closer to the top right indicate better ID accuracy and OoDD performance.}
    \label{Fig1: MOS, OPENOOD compare}
    \vspace{-10pt}
\end{figure*}

Recently, the emergence of vision-language models (VLMs) like CLIP~\cite{radford2021learning} has expanded OoDD research to include multi-modal approaches that leverage textual information~\cite{fort2021exploring, esmaeilpour2022zero,ming2022delving,wang2023clipn,miyai2024locoop,jiang2024negative,li2024learning}. These multi-modal approaches enable \textit{zero-shot} (ZS) OoDD without fine-tuning encoders. For example, 
MCM~\cite{ming2022delving} calculates an OoD score as the maximum value of the scaled softmax applied to the cosine similarities between an image embedding and the ID text embeddings.
NegLabel~\cite{jiang2024negative} not only uses ID texts but also additional textual information, negative labels distinct from ID texts, as auxiliary information for OoDD. 
While both methods utilize ID texts, NegLabel demonstrates superior OoDD performance compared to MCM.
This enhanced performance is due to its ability to fully leverage textual information and effectively utilize the knowledge embedded in pretrained VLMs.
Both methods are frequently employed as post-hoc OoD scoring functions due to their efficiency and effectiveness as ZS approaches.
Despite the notable ZS OoDD performance, these approaches can be suboptimal for downstream datasets. As shown in Fig.~\ref{Fig1: b}, even when utilizing textual information, ZS methods underperform in ID accuracy compared to \textit{single-modal fine-tuning} (SMFT) methods (e.g., LP, FFT, LP-FT), which rely solely on visual input data. 
While ZS OoDD research has primarily focused on enhancing OoDD performance, it often overlooks the importance of maintaining high ID accuracy, which is crucial for the overall reliability of the model.

Only a few recent studies~\cite{ming2024howto,ICEIC2024} have focused on the impact of \textit{multi-modal fine-tuning} (MMFT) and \textit{prompt learning} (PL) from a comprehensive perspective. These studies have shown the potential of fine-tuning methods in enhancing both ID accuracy and OoDD performance. Based on these findings, 
we explore the potential performance improvements achievable by integrating MMFT or PL methods with effective OoD scoring functions.
However, our experiments and analysis (see Sections~\ref{sec:Experimental results} and~\ref{sec:Analysis of Hyper}) reveal that existing fine-tuning methods are incapable of effectively leveraging the pretrained textual knowledge embedded in VLMs, despite showing significantly better ID accuracy compared to ZS methods.
We examine the underlying cause of this limitation and attribute it to the modality gap~\cite{liang2022mind, oh2024geodesic, shi2023towards} within the ID embeddings in the multi-modal representation space, where image and text embeddings are unexpectedly separated on a hypersphere as illustrated in Figs.~\ref{Fig1: a} (top) and~\ref{DOSNES ZS}.
The impact of the modality gap is minimal for classification tasks because selecting the highest similarity is relatively straightforward as shown in the upper right of Fig.~\ref{Fig1: a}. However, when we fully leverage the textual information for OoDD, this gap may hinder the distinction between ID texts and negative concept labels, potentially leading to a decrease in OoDD performance.
We discuss this challenge further in Section~\ref{sec:Analysis of Hyper}.

To tackle this issue, we propose a novel MMFT method using a regularization term called cross-modal alignment (CMA). Our methodology aims to fine-tune the image and text encoders to improve both OoDD performance and ID accuracy, forming an embedding space similar to the bottom hypersphere in Fig.~\ref{Fig1: a}.
The key idea behind CMA involves two main components: 1) \textit{alignment} between the image and text modalities of the ID data to effectively separate negative text embeddings, and 2) \textit{correspondence} between matching ID image-text pairs to maintain ID accuracy. Our approach is designed to bring image-text pairs of ID data closer together while effectively separating ID images and texts from negative concepts. To implement these ideas, we employ a contrastive loss for each modality with an additional CMA regularization loss. We show that our method is equivalent to modeling and optimizing a joint energy-based model with a generative term in the hyperspherical embedding space, as detailed in Section~\ref{sec:Methods}.

Using the widely adopted MOS benchmark~\cite{huang2021mos} for VLM-based OoDD and the recent OpenOOD v1.5 benchmark datasets~\cite{zhang2023openood}, which considers both Near-OoD and Far-OoD scenarios, we demonstrate that our method achieves superior performance in the ImageNet-1k dataset OoD setting. We also analyze the impact of reducing the ID modality gap on the hypersphere from the perspectives of uniformity and alignment. Our main contributions are summarized as follows:

\begin{itemize}

\item We introduce a novel MMFT method called cross-modal alignment (CMA), which effectively aligns image-text embeddings on a hypersphere. This alignment enhances both ID accuracy and OoDD performance by leveraging the pretrained textual knowledge in VLMs.

\item We demonstrate that minimizing our objective is equivalent to maximizing the log-likelihood of a joint energy-based model on a hyperspherical representation space.

\item Our approach shows state-of-the-art performance on the MOS and standard OpenOOD v1.5 benchmarks. Also, we provide an in-depth analysis of how the hyperspherical embedding space built by CMA enhances OoDD.

\end{itemize}

\section{Related works}
\label{sec:Related works}
Out-of-distribution detection (OoDD) enables models to distinguish between ID and OoD samples while maintaining accuracy on ID data. 
A variety of OoDD techniques have been proposed for visual inputs~\cite{ hsu2020generalized, liu2020energy, kirichenko2020normalizing, hendrycks2022scaling, sun2022out, tao2022non, wang2022vim, miyai2023can, ming2022CIDER, du2022siren, koo2024goe, kim2025reguide, kim2023unified}.
Recently, OoDD research has entered a new paradigm with the emergence of VLMs, such as CLIP~\cite{radford2021learning}, which leverage the textual information for OoDD~\cite{fort2021exploring, ming2022delving, zhou2022learning, zhou2022conditional, wang2023clipn, park2024powerfulness, jiang2024negative}.

Among these, several ZS OoDD methods~\cite{fort2021exploring, ming2022delving,esmaeilpour2022zero,wang2023clipn,jiang2024negative} have been introduced.
ZOC~\cite{esmaeilpour2022zero} performs OoDD by generating OoD candidate labels using a text description generator, while CLIPN~\cite{wang2023clipn} utilizes an additional text encoder capable of interpreting negative prompts. 
Certain ZS OoDD methods, such as MCM~\cite{ming2022delving} and NegLabel~\cite{jiang2024negative}, have demonstrated remarkable performance as post-hoc OoDD scoring functions.
Another approach, PL~\cite{miyai2024locoop, li2024learning}, involves training only a single or a few layers to learn prompts suitable for the task.
LoCoOp~\cite{miyai2024locoop} employs an OoD regularization technique that treats segments of an image's local features, such as ID-irrelevant elements like backgrounds, as OoD features during the training process.
NegPrompt~\cite{li2024learning} develops a set of negative prompts, each representing a negative aspect of a given class label, to clearly delineate the boundaries between ID and OoD images.

Previous research has primarily focused on leveraging the capabilities of VLMs by using textual information.
However, the aforementioned methods, which either freeze the encoder, partially utilize it, or train additional layers, may be suboptimal in fully leveraging the potential of pretrained VLMs. Recent works~\cite{ming2024howto, ICEIC2024} have improved OoDD performance through MMFT and PL methods, such as FLYP~\cite{10205046} and CoOp~\cite{zhou2022learning}. Building on these studies, we investigate fine-tuning strategies to maximize the use of VLMs' knowledge. We emphasize the potential of MMFT, which directly engages with the multi-modal representation space through fine-tuning all parameters, in contrast to ZS or PL methods. We focus on reducing the modality gap~\cite{oh2024geodesic, liang2022mind, shi2023towards, jiang2023understanding, ma2024bridging}, which can hinder distinguishability between ID and OoD texts. While certain MMFT methods, such as $m^2$-mix~\cite{oh2024geodesic}, address this gap, they are mainly aimed at robust fine-tuning for downstream tasks. In contrast, our method, CMA, is specifically designed to reduce the modality gap in ID embeddings, enabling more effective utilization of VLMs' pretrained knowledge for OoDD.
\section{Methods}
\label{sec:Methods}

\subsection{Preliminaries and Motivation}

CLIP uses a contrastive learning approach to learn multi-modal representations in a hyperspherical space by leveraging large-scale image and text data. The contrastive learning objective in CLIP trains both an image encoder $f(\cdot)$ and a text encoder $g(\cdot)$ simultaneously to ensure that matching image-text pairs are close in a shared representation space, while non-matching pairs are distant. Given a batch consisting of $B$ pairs of images and their corresponding text descriptions (i.e., positive texts) $ \{(I_k, T_k)\}_{k=1}^B $, the pairs of embeddings $ \{(i_k, t_k)\}_{k=1}^B = \{(f(I_k), g(T_k))\}_{k=1}^B $ are computed. Note that the dot product $i_k \cdot t_k$ is the cosine similarity between $i_k$ and $t_k$ since $f(\cdot)$ and $g(\cdot)$ produce $L_2$-normalized unit vectors (i.e., projected onto a hypersphere). Therefore, CLIP loss can be formulated as follows:
\begin{equation}
\label{eq : clip loss}
\mathcal{L}_\mathrm{CLIP}={1\over{2B}}\sum^B_{k=1}(\mathcal{L}^k_\mathrm{image}+\mathcal{L}^k_\mathrm{text}),
\end{equation}
where
\begin{equation}
\label{eq : clip image,text loss}
\mathcal{L}^k_\mathrm{image}=\displaystyle{-\log{\exp((i_k\cdot t_k/\tau))\over{\sum^B_{j=1}}\exp((i_k\cdot t_j)/\tau)}},
\end{equation}
\begin{equation}
\mathcal{L}^k_\mathrm{text}=\displaystyle{-\log{\exp((i_k\cdot t_k/\tau))\over{\sum^B_{j=1}}\exp((i_j\cdot t_k)/\tau)}}.
\end{equation}\\
Here, $\tau$ is a temperature parameter that scales the similarities and is set to a learnable parameter in CLIP. The learned representations are commonly used for transfer learning, and recently, FLYP~\cite{10205046} has shown that employing the CLIP loss for fine-tuning is highly effective for both ID accuracy and OoD generalization.

The contrastive loss is fundamentally focused on discrimination. It is known that the numerator of this loss corresponds to the alignment term, while the denominator corresponds to the uniformity term~\cite{wang2020understanding,wang2021understanding}. However, interpreting the denominator of the CLIP loss, which applies to multi-modal contrastive learning, as a uniformity term can be misleading. Specifically, $\mathcal{L}^k_\mathrm{image}$ can be decomposed as $\mathcal{L}^k_\mathrm{image}=-(i_k\cdot t_k)/\tau+\log\sum^B_{j=1}\exp((i_k \cdot t_j)/\tau)$. When an image is given, the first term increases the similarity with the positive text, while the second term decreases the similarity with all texts, including the positive text. Consequently, instead of uniformly distributing the text embeddings on the hypersphere, they form clusters distinct from the image embeddings. This tendency is further reinforced when the temperature parameter is sufficiently small, as even a slight increase in similarity with the positive text compared to other texts can significantly reduce the loss. The same applies when a text is given. Therefore, the CLIP loss primarily encourages contrastive learning across modalities rather than within each modality, leading to distinct clusters for each modality's embeddings as discussed in~\cite{liang2022mind} and~\cite{oh2024geodesic}.

Given that CLIP is trained on a vast amount of image and text data, it demonstrates remarkable OoDD performance even in a ZS setting~\cite{ming2022delving,wang2023clipn, jiang2024negative}.
Among the ZS methods leveraging CLIP’s rich multi-modal representations, NegLabel~\cite{jiang2024negative} is particularly effective. This approach identifies labels that are distant from ID texts and uses these negative labels for OoDD. Specifically, when an image is given, NegLabel determines whether the image is OoD by comparing the distances between the image and the ID texts with those between the image and the negative labels. Fine-tuning methods like FLYP enhance discrimination for ID classes, thereby achieving strong OoDD performance. However, they do not effectively utilize the pretrained negative concepts as NegLabel does. This limitation arises because these methods retain the embedding tendencies of the original CLIP loss, resulting in clustered embeddings. Consequently, the text embeddings still form clusters, which limits the improvement of separability from negative labels.

\subsection{Cross-Modal Alignment}
To effectively utilize pretrained multi-modal representations for OoDD through fine-tuning, it is essential to ensure that the image-text pairs corresponding to ID become closer, while also sufficiently separating the ID images and texts from negative concepts (i.e., negative labels).  To obtain representations that satisfy these properties, we propose a cross-modal alignment (CMA) loss for each modality that mitigates the separation effect between the image and text modalities of the ID data, which can be computed as:
\begin{align}
\mathcal{L}^k_\mathrm{image_{CMA}} &= -\log \sum^B_{j=1} \exp \left( i_k \cdot t_j / {\tau} \right), \\
\mathcal{L}^k_\mathrm{text_{CMA}} &= -\log \sum^B_{j=1} \exp \left( i_j \cdot t_k / {\tau} \right).
\end{align}
The proposed CMA losses work by globally increasing the similarity between ID image and text pairs, causing the ID modalities to be well-aligned on the hypersphere. As a result, the separability from the pretrained negative concepts can be enhanced, allowing for more effective OoDD. By introducing a hyperparameter $\lambda$ that controls the alignment strength, the final loss can be written as:
\begin{equation}
\mathcal{L}_\mathrm{CMA}=\mathcal{L}_\mathrm{CLIP} + 
{\lambda\over{2B}}\sum^B_{k=1}(\mathcal{L}^k_\mathrm{image_{CMA}}+\mathcal{L}^k_\mathrm{text_{CMA}}).
\label{eq:objective}
\end{equation}
Equivalently, $\mathcal{L}^k_\mathrm{image}$ in Eq.~\ref{eq : clip image,text loss} is revised to $-(i_k\cdot t_k)/\tau+(1-\lambda)\log\sum^B_{j=1}\exp((i_k \cdot t_j)/\tau)$, and $\mathcal{L}^k_\mathrm{text}$ is revised in the same manner.

\paragraph{OoD score.} To fully exploit the pretrained negative concepts, we follow an OoD scoring scheme of NegLabel. Given $C$ text embeddings of ID classes $\{t_k\}_{k=1}^C$ and $M$ negative text embeddings $\{\tilde{t}_k\}_{k=1}^M$ obtained by the NegMining algorithm~\cite{jiang2024negative}, the OoD score for an image $I$ is obtained by:
\begin{equation}
S(I) = \frac{\sum^C_{k=1}\exp\left({i \cdot t_k}/{\tau}\right)}{\sum^C_{k=1}\exp\left({i \cdot t_k}/{\tau}\right) + \sum^M_{k=1}\exp\left({i \cdot \tilde{t}_k}/{\tau}\right)},
\label{eq:ood score}
\end{equation}
where $i$ represents the image embedding of $I$ from the image encoder $f(\cdot)$. 

\paragraph{Relation to EBMs.} 
The proposed CMA objective is closely related to energy-based models (EBMs)~\cite{lecun2006tutorial}. We demonstrate that minimizing our objective in Eq.~\ref{eq:objective} is equivalent to maximizing the log-likelihood of a joint EBM in a hyperspherical embedding space. This equivalence allows us to extend the EBM framework to the multi-modal domain and demonstrates its effectiveness for OoDD tasks. Specifically, our CMA objective regulates the importance of the generative term (i.e., the log marginal density) in the EBM framework through the hyperparameter $\lambda$.

By defining the similarity between image embeddings $i$ and text embeddings $t$ as the negative energy $E_\theta(i,t) = - (i \cdot t)$, the joint distribution of images and texts can be expressed as a Gibbs-Boltzmann distribution as 
$q_\theta(i, t) = \exp(-E_\theta(i, t)/\tau)/{Z(\theta)}$
where $Z$ is the partition function, i.e., $Z(\theta) = \int_i \int_t \exp(-E_\theta(i, t)/\tau) \, \mathrm{d}t \, \mathrm{d}i$
which is generally intractable. Note that $\theta$ represents a learnable parameter set of $f(\cdot)$ and $g(\cdot)$. Given a data distribution $p$, $\theta$ can be estimated by maximizing the expected log-likelihood of $q_\theta$ under $p$:
\begin{align}
\label{eq 6}
\max_\theta \mathbb{E}_p \left[ \log q_\theta(i,t) \right] &= 
\max_\theta \frac{1}{2} \mathbb{E}_p \left[ \log q_\theta(t|i) + \log q_\theta(i) \right] \notag \\
& + \frac{1}{2} \mathbb{E}_p \left[ \log q_\theta(i|t) + \log q_\theta(t) \right].
\end{align}
Hereafter, we will consider only the first term on the right-hand side as the same analogy applies to the second term.

In EBMs, $\mathbb{E}_p \left[ \log q_\theta(t|i) \right]$ and $\mathbb{E}_p \left[ \log q_\theta(i) \right]$ are interpreted as a discriminative term and a generative term, respectively. Given samples from $p$, the discriminative term can be estimated by the contrastive loss function as described in~\cite{energy:kim2022energy}, and therefore, the empirical form of it corresponds to $-\mathcal{L}^k_\mathrm{image}$ in Eq.~\ref{eq : clip image,text loss}. Therefore, the contrastive loss can be equated to a constrained form of EBMs, which does not consider the generative term. The generative term can be written as $\mathbb{E}_p \left[ \log q_\theta(i) \right] = - \mathbb{E}_p[E_\theta(i)] - \log Z(\theta)$, and the empirical estimation is generally intractable due to $Z(\theta)$. However, we can ignore the partition function since our embeddings $i$ and $t$ are in the hyperspherical space.

The von Mises-Fisher (vMF) distribution is a probability distribution on the hypersphere in $\mathbb{R}^p$ for the $p$-dimensional unit vectors. Given $t$ be the mean direction of vMF, the probability density function for the image embeddings is defined as $
q_\theta(i|t) = \mathcal{C}_p({1}/{\tau})\exp\left({-E_\theta(i, t)}/{\tau}\right),
$ where $\mathcal{C}_p(1/{\tau})$ denotes the normalization factor. Thus, the log marginal density of $i$ and its empirical estimate can be written as:
\begin{align}
\label{eq:log-marginal-density}
\log q_\theta(i) &= \log \int_t q_\theta(i|t) q_\theta(t) \, \mathrm{d}t \\
&\approx \log \sum_{t \in B} q_\theta(i|t) \\
&\approx \log \sum_{t \in B} \exp\left({-E_\theta(i, t)}/{\tau}\right) + C
\end{align}
where a constant $C=\log \mathcal{C}_p({1}/{\tau})$.
It should be noted that the first term on the right-hand side corresponds to $-\mathcal{L}^k_\mathrm{image_{CMA}}$: maximizing $\log q_\theta(i)$ is equivalent to minimizing $\mathcal{L}^k_\mathrm{image_{CMA}}$. 
Consequently, while the contrastive loss focuses on solely discrimination (i.e., the log conditional likelihood $\log q_\theta(t|i)$), our method also considers generation (i.e., the log marginal density $\log q_\theta(i)$).

\begin{table*}[t]
\fontsize{9pt}{10pt}\selectfont
\setlength{\tabcolsep}{0.4mm}
\centering
\caption{Comparison of OoDD performance on the MOS benchmark dataset. $\uparrow$ and $\downarrow$ denote that higher and lower values are better, respectively. The highest values are highlighted in bold, and the second-highest values are underlined.}
\label{tab:table1}
% \resizebox{\textwidth}{!}{%
\small{
\begin{tabular}{lcccccccccc}
\toprule
& \multicolumn{2}{c}{iNaturalist} & \multicolumn{2}{c}{SUN} & \multicolumn{2}{c}{Places} & \multicolumn{2}{c}{Textures} & \multicolumn{2}{c}{Average} \\ \cline{2-11} 
\textbf{Methods}    & \scriptsize{FPR95$\downarrow$} & \scriptsize{AUROC$\uparrow$} & \scriptsize{FPR95$\downarrow$} & \scriptsize{AUROC$\uparrow$} & \scriptsize{FPR95$\downarrow$} & \scriptsize{AUROC$\uparrow$} & \scriptsize{FPR95$\downarrow$} & \scriptsize{AUROC$\uparrow$} & \scriptsize{FPR95$\downarrow$} & \scriptsize{AUROC$\uparrow$} \\ 
\midrule
\textit{Zero-Shot (ZS)} & & & & & & & & & & \\
MCM & 32.28 & 94.40 & 39.33 & 92.28 & 44.94 & 89.83 & 57.98 & 85.99 & 43.63 & 90.63 \\
CLIPN & 23.94 & 95.27 & 26.17 & 93.93 & 33.45 & 92.28 & 40.83 & 90.93 & 31.10 & 93.10 \\
NegLabel & \underline{2.30} & 99.37 & 23.23 & \underline{95.14} & 39.85 & 90.98 & 46.49 & 89.64 & 27.97 & 93.78 \\
\midrule
\textit{Prompt Learning (PL)} & & & & & & & & & & \\
CoOp$\mbox{\scriptsize{MCM}}$ & 26.37 & 94.49 & 35.23 & 92.59 & 43.29 & 89.67 & 41.47 & 90.62 & 36.59 & 91.84 \\
CoOp$\mbox{\scriptsize{NegLabel}}$ & 4.95 & 98.90 & 25.76 & 94.59 & \underline{30.07} & \underline{93.33} & 44.35 & 89.59 & 26.28 & \underline{94.10} \\
LoCoOp$\mbox{\scriptsize{MCM}}$ & 23.08 & 95.46 & 33.39 & 93.25 & 40.74 & 90.52 & 40.75 & \underline{91.14} & 34.49 & 92.59 \\
LoCoOp$\mbox{\scriptsize{NegLabel}}$ & 3.19 & 99.25 & 46.63 & 90.58 & 55.44 & 87.65 & 46.03 & 89.85 & 37.83 & 91.83 \\
\midrule
\textit{Single-modal Fine-tuning (SMFT)} & & & & & & & & & & \\
LP$\mbox{\scriptsize{MSP}}$ & 41.61 & 90.87 & 67.03 & 81.07 & 66.03 & 81.25 & 62.82 & 81.46 & 59.37 & 83.66 \\
LP$\mbox{\scriptsize{ODIN}}$ & 28.46 & 94.90 & 51.13 & 89.17 & 51.13 & 88.39 & 53.83 & 86.40 & 46.14 & 89.72 \\
LP$\mbox{\scriptsize{Energy}}$ & 31.09 & 94.48 & 58.42 & 86.95 & 65.90 & 83.68 & 75.48 & 78.23 & 57.72 & 85.83 \\
FFT$\mbox{\scriptsize{MSP}}$ & 51.66 & 87.87 & 70.55 & 79.86 & 71.77 & 79.42 & 70.82 & 78.42 & 66.20 & 81.39 \\
FFT$\mbox{\scriptsize{ODIN}}$ & 19.05 & 96.38 & 39.24 & 90.75 & 46.93 & 88.52 & 52.89 & 85.93 & 39.53 & 90.40 \\
FFT$\mbox{\scriptsize{Energy}}$ & 24.46 & 95.62 & 58.98 & 84.77 & 59.92 & 83.89 & 55.60 & 84.45 & 49.74 & 87.18 \\
LP-FT$\mbox{\scriptsize{MSP}}$ & 44.02 & 89.81 & 66.60 & 80.78 & 69.20 & 80.06 & 65.59 & 79.44 & 61.35 & 82.52 \\
LP-FT$\mbox{\scriptsize{ODIN}}$ & 27.22 & 94.28 & 46.65 & 87.80 & 52.26 & 86.00 & 53.58 & 84.18 & 44.93 & 88.06 \\
LP-FT$\mbox{\scriptsize{Energy}}$ & 29.16 & 93.93 & 48.27 & 87.47 & 53.24 & 85.68 & 54.57 & 83.81 & 46.31 & 87.72 \\
\midrule
\textit{Multi-modal Fine-tuning (MMFT)} & & & & & & & & & & \\
FLYP$\mbox{\scriptsize{MCM}}$ & 24.86 & 94.35 & 39.81 & 90.58 & 47.92 & 87.16 & 41.19 & 89.34 & 38.44 & 90.36 \\
FLYP$\mbox{\scriptsize{NegLabel}}$ & 3.16 & 99.31 & 23.48 & 94.82 & 37.23 & 90.86 & 41.70 & 89.27 & 26.39 & 93.57 \\
$m^2$-mix$\mbox{\scriptsize{MCM}}$& 22.41 & 95.61 & 39.18 & 91.85 & 47.07 & 88.72 & 43.44 & 90.13 & 38.02 & 91.58\\

$m^2$-mix$\mbox{\scriptsize{NegLabel}}$& 2.39 & \underline{99.43} & \underline{23.03} & 94.86 & 35.55 & 91.21 & \underline{36.65} & 90.68 & \underline{24.40} & 94.05 \\
\rowcolor[gray]{0.9}CMA$\mbox{\scriptsize{MCM}}$ & 22.95 & 95.65 & 40.01 & 91.78 & 48.83 & 88.41 & 44.93 & 89.87 & 39.18 & 91.43 \\
\rowcolor[gray]{0.9}CMA$\mbox{\scriptsize{NegLabel}}$& \textbf{1.65} & \textbf{99.62} & \textbf{16.84} & \textbf{96.36} & \textbf{27.65} & \textbf{93.11} & \textbf{33.58} & \textbf{91.64} & \textbf{19.93} & \textbf{95.13} \\
\bottomrule

\end{tabular}%
% }
}
\vspace{-10pt}
\end{table*}
\section{Experiments}
\label{sec:Experiments}

\subsection{Experimental Setup}

\paragraph{Datasets.} We evaluate our method in comparison to various previous approaches using the ImageNet-1k OoDD benchmarks, MOS~\cite{huang2021mos}, and OpenOOD v1.5~\cite{zhang2023openood}. Following the MOS, we use ImageNet-1k as the ID dataset along with semantically diverse OoD datasets such as iNaturalist~\cite{van2018inaturalist}, SUN~\cite{xiao2010sun}, Places~\cite{zhou2017places}, and Textures~\cite{cimpoi2014describing}. Additionally, we employ the standard OpenOOD v1.5 benchmarks to assess performance in both Near- and Far-OoD scenarios, also using ImageNet-1k as the ID dataset. For Near-OoD, datasets such as SSB-hard~\cite{vaze2021open} and NINCO~\cite{bitterwolf2023or}, which are semantically more challenging to distinguish from the ID dataset, are included. For Far-OOD, we utilize datasets such as iNaturalist, Textures, and OpenImage-O~\cite{kuznetsova2020open}. 

\vspace{-10pt}
\paragraph{Baselines.} We compare the performance of our method for OoDD against baselines categorized into four main training strategies\footnote{For SMFT and MMFT, results are obtained using the entire dataset. For PL, results are reproduced using the default 16-shot setting. Performance across various shot settings is reported in Appendix~\ref{abl:coop locoop}.}: (1) ZS, (2) PL, (3) SMFT, and (4) MMFT. 
For a comprehensive assessment, we compare both OoDD performance and ID accuracy.

For ZS methods, we employ MCM~\cite{ming2022delving}, NegLabel~\cite{jiang2024negative} and CLIPN~\cite{wang2023clipn}. 
% These methods can be generally used as OoD scoring functions across various VLMs. 
In PL, we consider CoOp~\cite{zhou2022learning} and LoCoOp~\cite{miyai2024locoop}. SMFT methods, a common approach in visual transfer learning, involve training the pretrained visual encoder on the ID dataset alongside a classification layer. We employ linear-probing (LP), full-fine-tuning (FFT), and linear-probing then fine-tuning (LP-FT)~\cite{kumar2021fine}. For MMFT, we use FLYP~\cite{10205046} and $m^2\text{-mix}$~\cite{oh2024geodesic}. FLYP mimics the CLIP pre-training scheme for fine-tuning and has been proven to perform well in both classification and OoD generalization~\cite{10205046}. $m^2\text{-mix}$ aims to mitigate the modality gap on the hypersphere by utilizing hard negative embeddings obtained through the mixing of image and text embeddings.

\begin{table*}[t]
\centering
\caption{Comparison of OoDD performance on the OpenOOD v1.5 benchmark dataset. The terms Near-OoD and Far-OoD refer to the average OoDD performance across datasets in each respective scenario.}
\label{tab:OpenOOD v1.5}
% \resizebox{\textwidth}{!}{%
\fontsize{9pt}{10pt}\selectfont
\setlength{\tabcolsep}{0.1mm}
\begin{tabular}{lcccccccccccccc}
\toprule
          & \multicolumn{2}{c}{SSB-hard} & \multicolumn{2}{c}{NINCO}& \multicolumn{2}{c}{Near-OoD} & \multicolumn{2}{c}{iNaturalist} & \multicolumn{2}{c}{Textures} & \multicolumn{2}{c}{Openimage-O} & \multicolumn{2}{c}{Far-OoD} \\ \cline{2-15} 
\textbf{Methods}    & \scriptsize{FPR95$\downarrow$} & \scriptsize{AUROC$\uparrow$} & \scriptsize{FPR95$\downarrow$} & \scriptsize{AUROC$\uparrow$} & \scriptsize{FPR95$\downarrow$} & \scriptsize{AUROC$\uparrow$} & \scriptsize{FPR95$\downarrow$} & \scriptsize{AUROC$\uparrow$} & \scriptsize{FPR95$\downarrow$} & \scriptsize{AUROC$\uparrow$} & \scriptsize{FPR95$\downarrow$} & \scriptsize{AUROC$\uparrow$} & \scriptsize{FPR95$\downarrow$} & \scriptsize{AUROC$\uparrow$} \\ 
\midrule
\textit{Zero-Shot (ZS)} & & & & & & & & & & & & & & \\

MCM & 88.32 & 65.05 & 79.23 & 74.44 & 83.78 & 69.74 & 32.25 & 94.41 & 55.62 & 87.11 & 44.34 & 91.23 & 44.07 & 90.92 \\

NegLabel & 81.87 & 71.32 & 69.82 & 77.09 & 75.85 & 74.21 & \underline{2.32} & 99.36 & 44.98 & 90.56 & 31.10 & 93.10 & 26.14 & 94.34 \\
\midrule

\textit{Prompt Learning (PL)} & & & & & & & & & & & & & & \\

CoOp$\mbox{\scriptsize{MCM}}$ & 86.10 & 67.72 & 77.24 & 74.59 & 81.67 & 71.16 & 26.31 & 94.50 & 38.38 & 91.92 & 37.64 & 92.08 & 34.11 & 92.83 \\

CoOp$\mbox{\scriptsize{NegLabel}}$ & 68.20 & 78.72 & 57.71 & 84.20 & 62.96 & 81.46 & 4.96 & 98.90 & 42.74 & 90.28 & 23.18 & 95.25 & 23.63 & 94.81 \\

LoCoOp$\mbox{\scriptsize{MCM}}$ & 87.38 & 66.23 & 77.04 & 73.46 & 82.21 & 69.84 & 22.98 & 95.48 & 37.89 & 92.27 & 37.44 & 92.09 & 32.77 & 93.28 \\

LoCoOp$\mbox{\scriptsize{NegLabel}}$ & 76.03 & 74.42 & 66.51 & 80.96 & 71.27 & 77.69 & 3.70 & 99.16 & 43.87 & 90.59 & 29.29 & 93.65 & 25.62 & 94.47 \\
\midrule

\textit{Single-modal Fine-tuning (SMFT)} & & & & & & & & & & & & & & \\

LP$\mbox{\scriptsize{MSP}}$ & 85.77 & 68.15 & 73.94 & 79.64 & 79.86 & 73.89 & 41.37 & 90.90 & 60.02 & 84.11 & 52.91 & 87.14 & 51.43 & 87.38 \\

LP$\mbox{\scriptsize{ODIN}}$ & 85.15 & 71.79 & 75.08 & 81.91 & 80.12 & 76.85 & 28.45 & 94.92 & 50.64 & 88.68 & 41.93 & 91.86 & 40.34 & 91.82 \\

LP$\mbox{\scriptsize{Energy}}$ & 87.93 & 68.80 & 79.40 & 78.14 & 83.67 & 73.47 & 31.09 & 94.48 & 73.84 & 80.60 & 58.20 & 88.08 & 54.38 & 87.72 \\

FFT$\mbox{\scriptsize{MSP}}$ & 81.00 & 72.77 & 69.43 & 81.54 & 75.21 & 77.16 & 51.66 & 87.85 & 68.74 & 80.46 & 56.45 & 86.31 & 58.95 & 84.88 \\

FFT$\mbox{\scriptsize{ODIN}}$ & 65.15 & \underline{82.94} & 56.71 & 87.16 & 60.93 & \underline{85.05} & 19.06 & 96.37 & 49.88 & 87.53 & 25.18 & 94.83 & 31.37 & 92.91 \\

FFT$\mbox{\scriptsize{Energy}}$ & \textbf{54.21} & \textbf{85.91} & 55.47 & \underline{87.18} & \textbf{54.84} & \textbf{86.54} & 24.55 & 95.62 & 52.97 & 86.41 & 29.32 & 93.95 & 35.61 & 91.99 \\

LP-FT$\mbox{\scriptsize{MSP}}$ & 80.02 & 72.56 & 69.16 & 80.78 & 74.59 & 76.67 & 43.96 & 89.82 & 63.24 & 81.49 & 50.43 & 87.77 & 52.54 & 86.36 \\

LP-FT$\mbox{\scriptsize{ODIN}}$ & 73.52 & 77.83 & 64.13 & 82.38 & 68.82 & 80.10 & 27.34 & 94.28 & 50.85 & 86.34 & 32.84 & 93.05 & 37.01 & 91.22 \\

LP-FT$\mbox{\scriptsize{Energy}}$ & 75.81 & 76.48 & 65.08 & 82.05 & 70.44 & 79.27 & 29.22 & 93.93 & 51.84 & 86.04 & 34.36 & 92.85 & 38.47 & 90.94 \\
\midrule

\textit{Multi-modal Fine-tuning (MMFT)} & & & & & & & & & & & & & & \\

FLYP$\mbox{\scriptsize{MCM}}$ & 73.89 & 75.13 & 61.88 & 79.93 & 67.89 & 77.53 & 24.73 & 94.36 & 37.48 & 90.42 & 29.66 & 93.20 & 30.62 & 92.66 \\

FLYP$\mbox{\scriptsize{NegLabel}}$ & 67.56 & 80.59 & 59.76 & 83.33 & 63.66 & 81.96 & 3.14 & 99.31 & 37.77 & 91.04 & 22.87 & 94.22 & 21.26 & 94.86 \\

$m^2$-mix$\mbox{\scriptsize{MCM}}$& 77.63 & 77.78 & 66.17 & 83.43 & 71.90 & 80.60 & 22.94 & 95.80 & 43.10 & 90.85 & 29.63 & 94.27 & 31.89 & 93.64\\

$m^2$-mix$\mbox{\scriptsize{NegLabel}}$& 66.57 & 79.72 &\underline{54.12} &84.93 &60.35 &82.32 & 2.34 & \underline{99.43} &\underline{32.48} &\underline{92.33} &\underline{17.80} &\underline{95.64} &\underline{17.54} &\underline{95.80} \\

\rowcolor[gray]{0.9}CMA$\mbox{\scriptsize{MCM}}$ & 78.28 & 74.95 & 65.64 & 82.18 & 71.96 & 78.56 & 23.03 & 95.65 & 41.47 & 91.18 & 29.25 & 94.35 & 31.25 & 93.72 \\

\rowcolor[gray]{0.9}CMA$\mbox{\scriptsize{NegLabel}}$ & \underline{62.80} & 81.71 & \textbf{49.70} & \textbf{87.21} & \underline{56.25} & 84.46 & \textbf{1.65} & \textbf{99.62} & \textbf{29.55} & \textbf{93.21} & \textbf{14.65} & \textbf{96.57} & \textbf{15.29} & \textbf{96.47} \\

\bottomrule
\end{tabular}%
% }
\vspace{-5pt}
\end{table*}

\begin{table}
\fontsize{9pt}{10pt}\selectfont
\setlength{\tabcolsep}{0.4mm}
\centering
\small 
\caption{ID Accuracy on ImageNet-1k}
\begin{tabular}{lc}
\toprule
\textbf{Methods}      & ID Acc. \\ 
\midrule
\textit{Zero-Shot (ZS)}  \\
MCM \& NegLabel & 66.60 \\ 
\midrule
\textit{Prompt Learning (PL)}  \\
CoOp & 71.95\\
LoCoOp & 71.72 \\
\midrule
\textit{Single-modal Fine-tuning (SMFT)}  \\ 
LP & 79.22 \\
FFT & 81.44 \\
LP-FT & 81.48 \\
\midrule
\textit{Multi-modal Fine-tuning (MMFT)}  \\  
FLYP & 82.58 \\
$m^2$-mix & \textbf{82.67}\\

\rowcolor[gray]{0.9}CMA & \underline{82.64} \\
\bottomrule
\end{tabular}
\label{tab:ID accuracy}
\vspace{-10pt}
\end{table}

\vspace{-10pt}
\paragraph{OoD scoring functions.} 
For PL and MMFT, we utilize both MCM and NegLabel\footnote{Using the same NegMining approach as NegLabel. The results reported in Tables~\ref{tab:table1} and \ref{tab:OpenOOD v1.5} are based on experiments conducted without the grouping strategy, which does not affect the performance superiority. For more details, see Appendix~\ref{abl:grouping strategy}.} as OoD scores. For SMFT, which cannot utilize textual information, we adopt different OoD scoring methods. In this case, we employ commonly used post-hoc methods that serve as standard baselines in visual OoDD research, such as MSP~\cite{hendrycks2016baseline}, ODIN~\cite{liang2018enhancing}, and Energy~\cite{liu2020energy}. 

\vspace{-10pt}
\paragraph{Evaluation metrics.} For evaluation, we use the following metrics to provide a comprehensive assessment of performance: (1) the false positive rate of OoD images when the true positive rate of in-distribution images is at 95\% (FPR95), (2) the area under the receiver operating characteristic curve (AUROC), and (3) ID accuracy.

\vspace{-10pt}
\paragraph{Implementation details.} For our experiments\footnote{Our code is available at \href{https://github.com/ma-kjh/CMA}{https://github.com/ma-kjh/CMA-OoDD}.}, we use the pretrained CLIP-B/16 model, which is widely adopted in prior research~\cite{ming2022delving, jiang2024negative, wang2023clipn, zhou2022learning, miyai2024locoop}. In our method, we perform hyperparameter sweeps over learning rates from the set \{1e-4, 1e-5, 1e-6\} and the hyperparameter $\lambda$ for alignment strengths \{1e-1, 1e-2, 1e-3\}, using a batch size of 512.\footnote{Ablation studies on various prompts and the hyperparameter $\lambda$ are reported in Appendices~\ref{abl:prompt change} and~\ref{sup:lambda}. The performance superiority remains unchanged regardless of the prompts used.} These settings are based on~\citet{10205046}. We apply early stopping based on the accuracy of the ID validation set. Additional experimental details can be found in Appendix~\ref{sup:experimental settings app}.

\subsection{Experimental Results}
\label{sec:Experimental results}

Tables \ref{tab:table1} and \ref{tab:OpenOOD v1.5} show the FPR95 and AUROC performances on the MOS and OpenOOD v1.5 benchmark datasets, respectively, while Table \ref{tab:ID accuracy} presents the ID accuracy of each method. These results demonstrate the effectiveness of our proposed method in both OoDD and ID classification.

We observe that SMFT methods, which rely solely on the visual encoder, are less effective for OoDD compared to other approaches leveraging textual information. When utilizing multi-modal foundation models like CLIP, it becomes clear that textual information plays a crucial role in OoDD.

For the MOS and OpenOoD v1.5 Far-OoD benchmarks in Tables~\ref{tab:table1} and \ref{tab:OpenOOD v1.5}, ZS NegLabel improves FPR95 by 15.66\%, 17.93\% and AUROC by 3.15\%, 3.42\% over ZS MCM.
Similarly, both PL and MMFT benefit significantly from NegLabel.
Notably, our proposed method surpasses the existing state-of-the-art performance, achieving an FPR95 of 19.93\% and an AUROC of 95.13\% on the MOS benchmark. 
Our method also demonstrates strong performance compared to baselines for the OpenOOD v1.5 Near- and Far-OoD scenarios, as shown in Table~\ref{tab:OpenOOD v1.5}, though it shows only moderate results on the SSB-hard dataset. In this case, $\text{FFT}_{\text{Energy}}$ achieves better results than other methods. This observation suggests that while OoDD methods leveraging textual information are generally effective in Far-OoD settings, the advantage becomes less pronounced in semantically similar datasets, e.g., Near-OoD scenarios.\footnote{To further investigate the behavior in Near-OoD scenarios, we perform additional experiments on ImageNet-1k splits~\cite{palechor2023large}, as described in Appendix~\ref{sup:Imagenet-split}.} 
Moreover, the results of $m^2\text{-mix}$ in Tables~\ref{tab:table1} and~\ref{tab:OpenOOD v1.5} indicate that mitigating the modality gap improves OoDD performance by effectively leveraging the pretrained knowledge. In particular, our CMA shows that reducing the modality gap within ID embeddings is more effective for enhancing the distinguishability between ID and OoD texts.

While each method using textual information shows performance improvements when provided with negative texts (i.e., NegLabel) rather than relying solely on ID texts, the PL and MMFT methods do not consistently deliver superior OoDD performance compared to ZS. In Table~\ref{tab:table1}, the results reveal that $\text{CoOp}_{\text{NegLabel}}$ outperforms ZS NegLabel on the Places and Textures datasets, yet demonstrates inferior performance on the iNaturalist and SUN datasets. LoCoOp$\mbox{\scriptsize{NegLabel}}$ also shows much lower performance compared to ZS NegLabel on the SUN and Places datasets, despite being specialized for OoDD. For $\text{FLYP}_{\text{NegLabel}}$, we observe that its AUROC performance is lower than that of ZS NegLabel across all datasets, trailing by an average of 0.21\%. These results suggest that existing PL or MMFT methods do not enhance the distinction between ID and negative texts as effectively as ZS NegLabel. This observation demonstrates that not considering the use of pretrained knowledge can lead to performance degradation, even when the methods are fine-tuned for specific downstream datasets.
In contrast, our method $\text{CMA}_{\text{NegLabel}}$ effectively leverages these negative concepts to perform OoDD, which we further analyze from the perspective of modality gap using various metrics in Section~\ref{sec:Analysis of Hyper}.

Table \ref{tab:ID accuracy} presents ID accuracy on ImageNet-1k, demonstrating that our proposed CMA performs well in both OoDD and ID classification, exceeding the performance of many other methods.
Our results indicate that fine-tuning methods such as SMFT and MMFT achieve significantly better ID accuracy compared to ZS or PL methods. Notably, Tables~\ref{tab:table1} and \ref{tab:OpenOOD v1.5} highlight the advantages of utilizing textual information for OoDD, with ZS, PL, and MMFT outperforming SMFT. Therefore, fine-tuning the encoders is effective for improving ID accuracy, while leveraging textual information is particularly beneficial for OoDD.
Consequently, our proposed fine-tuning method, CMA, designed to fully leverage pretrained textual knowledge, achieves superior overall performance in both ID classification and OoDD.

\section{Analysis of Hyperspherical Embeddings}
\label{sec:Analysis of Hyper}

\begin{figure}[htp]
    \centering
    \begin{subfigure}{.24\linewidth}
        \centering
        \includegraphics[width=\linewidth]{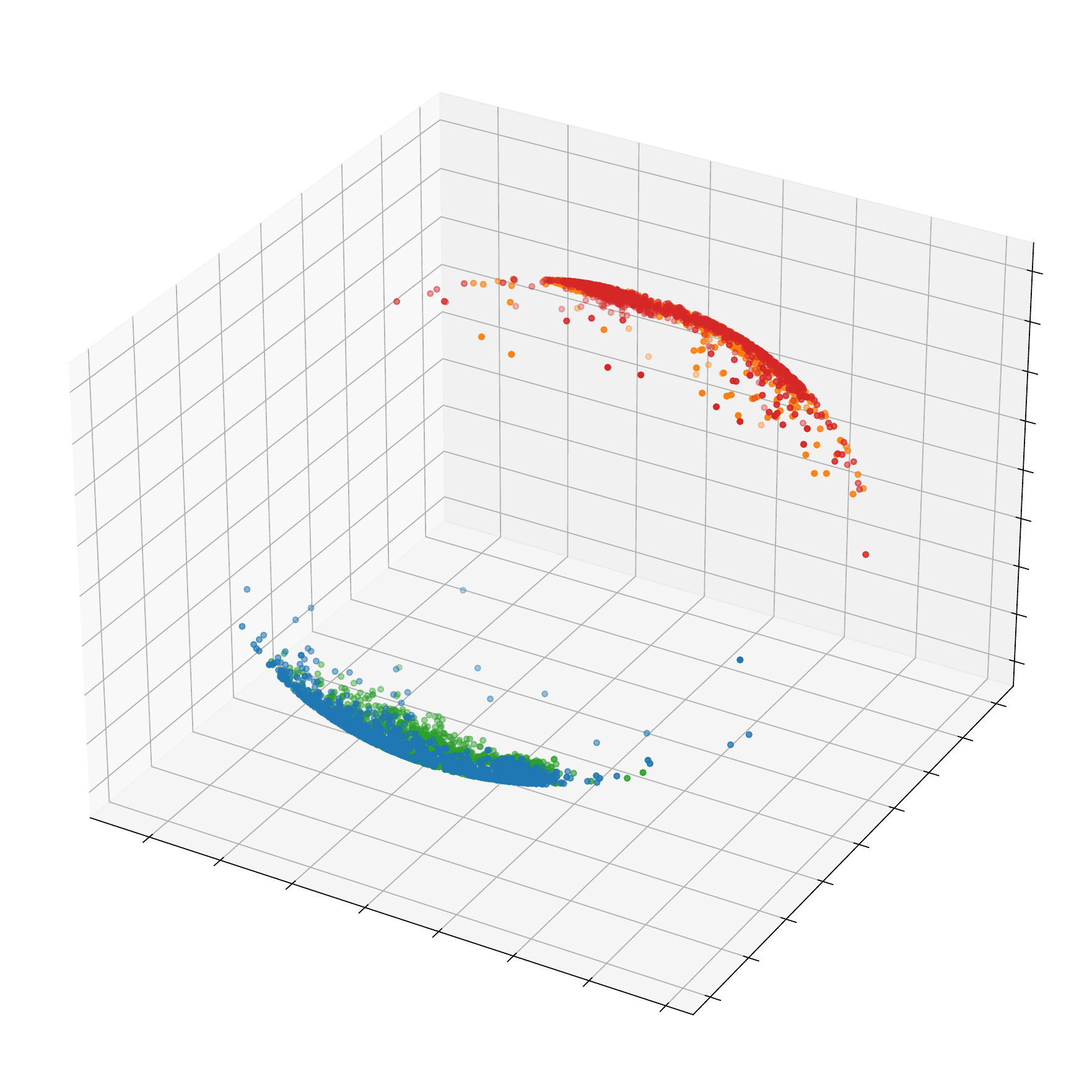}
        \caption{Zero-Shot}
        \label{DOSNES ZS}
    \end{subfigure}%
    % \hspace{-5mm} % Adjust this value to reduce the gap
    \begin{subfigure}{.24\linewidth}
        \centering
        \includegraphics[width=\linewidth]{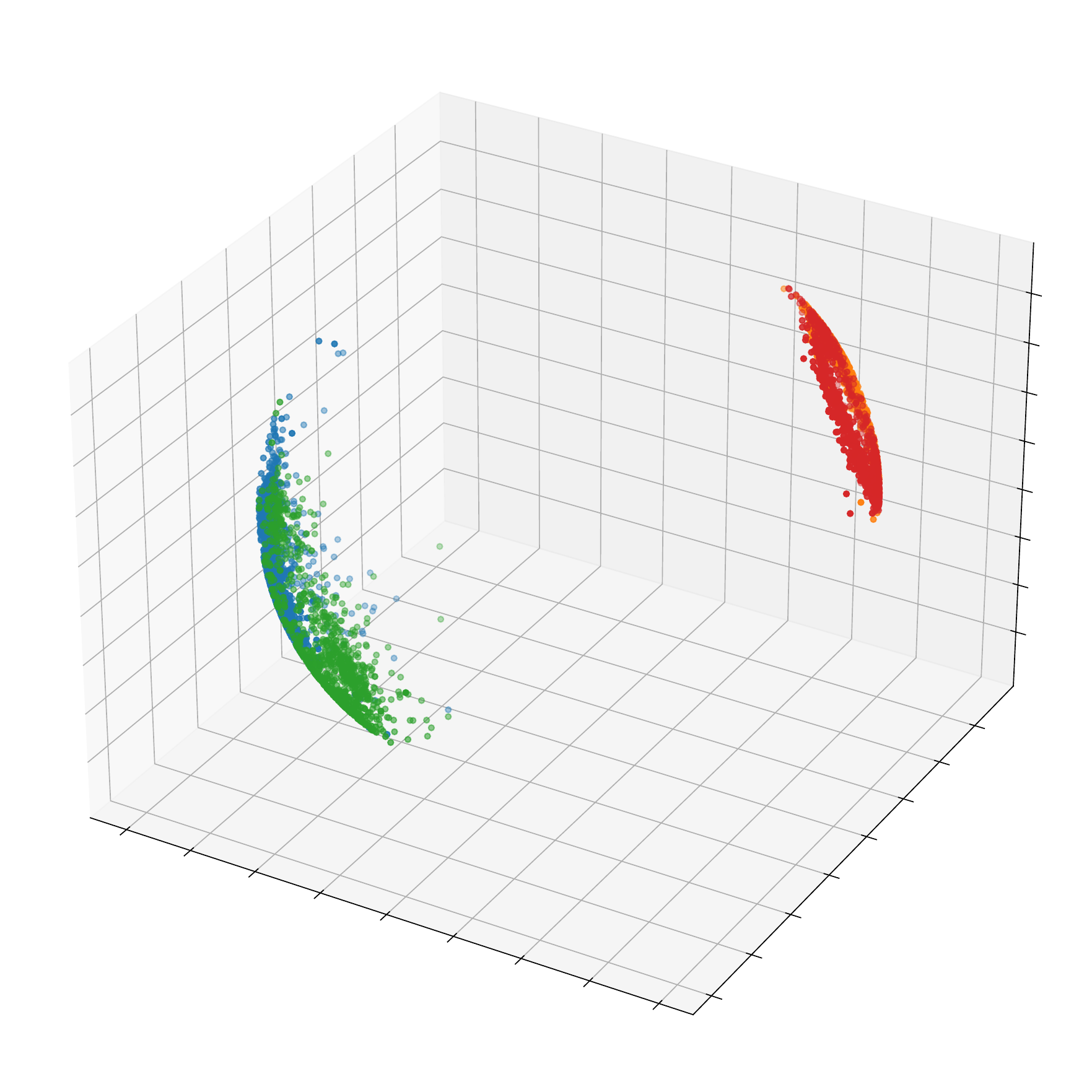}
        \caption{FLYP}
        \label{DOSNES FLYP}
    \end{subfigure}
    \begin{subfigure}{.24\linewidth}
        \centering
        \includegraphics[width=\linewidth]{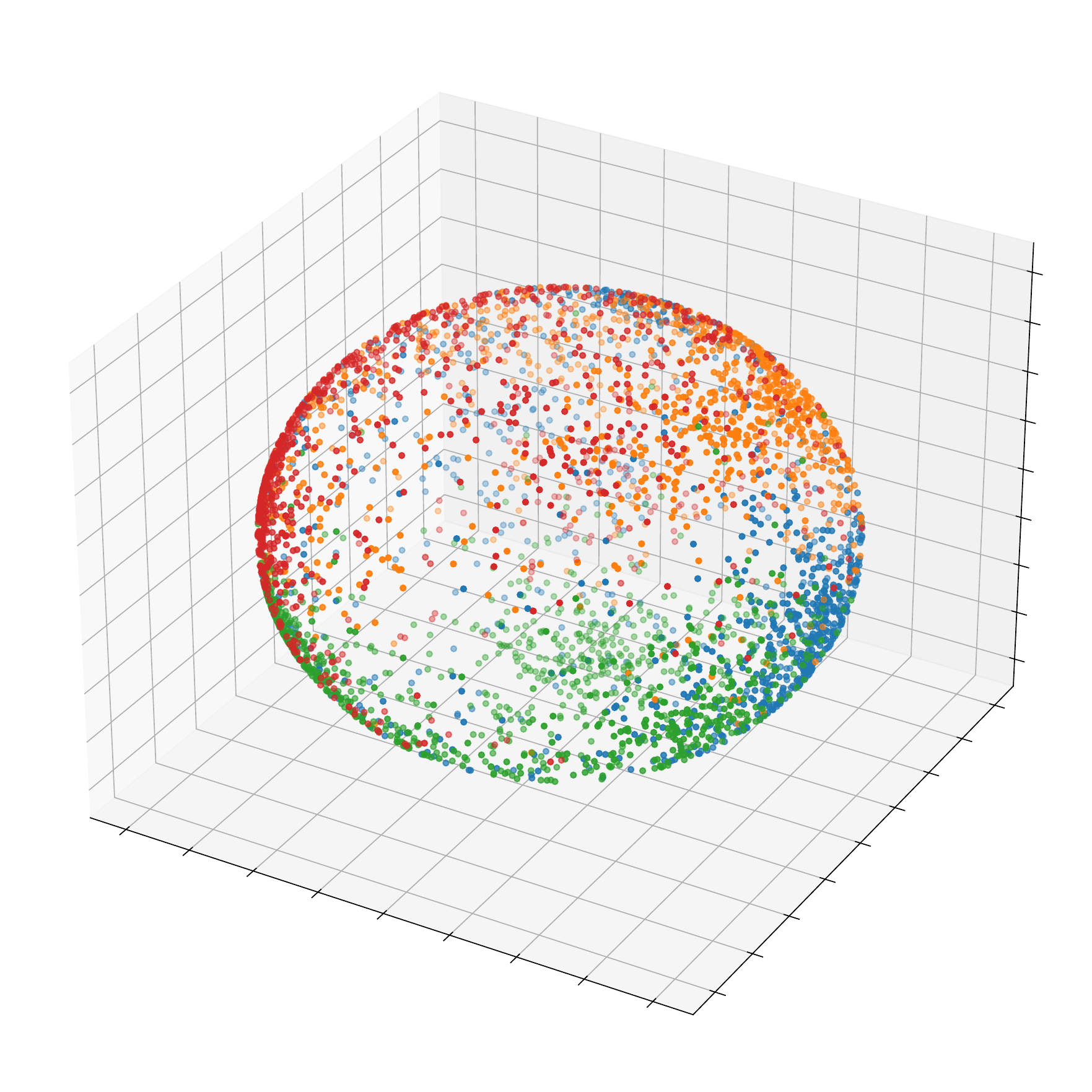}
        \caption{$m^2$-mix}
        \label{DOSNES mmmix}
    \end{subfigure}
    \hspace{-2mm} % Adjust this value to reduce the gap
    \begin{subfigure}{.24\linewidth}
        \centering
        \includegraphics[width=\linewidth]{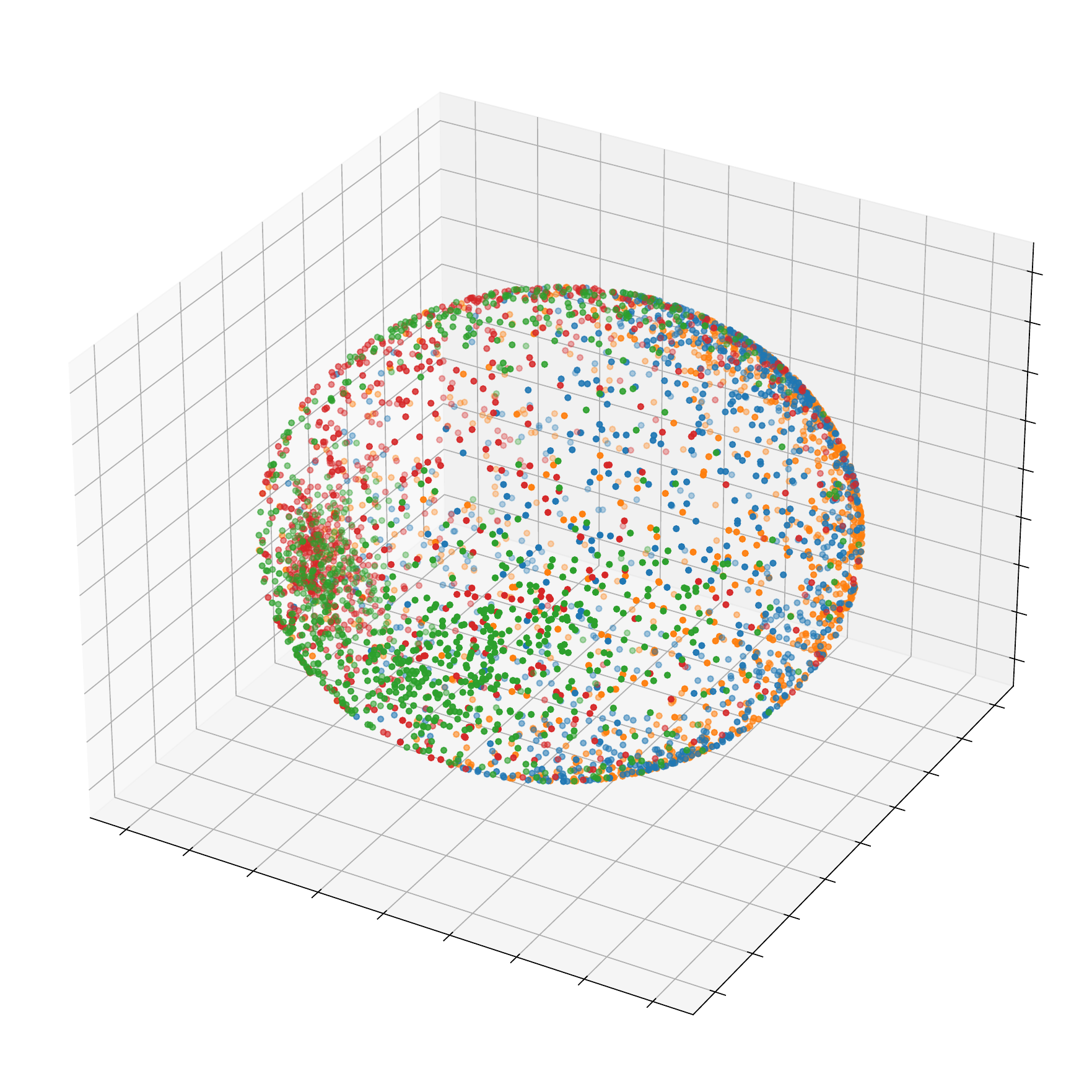}
        \caption{CMA}
        \label{DOSNES Ours}
    \end{subfigure}
    \vspace{-5pt}
    \caption{Visualization of DOSNES~\cite{lu2019doubly} on ImageNet-1k validation dataset and the MOS benchmark dataset. Blue and orange represent ID image and ID text embeddings, respectively, while green and red represent OoD image and OoD text embeddings. Additional visualizations are provided in the Appendix~\ref{sup:vis}.
    }
    \vspace{-10pt}
    \label{fig:DOSNES}
\end{figure}

In this section, we aim to provide insights into how the embedding space shapes and leverages the pretrained knowledge of VLMs to achieve effective performance in OoDD.
As shown in Figs.~\ref{Fig1: a} (top) and~\ref{DOSNES ZS}, the embedding space of CLIP is bipartite, with images and texts being separated. This phenomenon is referred to as the modality gap~\cite{liang2022mind,oh2024geodesic}.
It is important to note that effectively utilizing textual information for OoDD requires ID images and ID texts to be closer together in the embedding space than OoD texts. However, the modality gap can hinder this ideal scenario. Reducing the modality gap between ID images and ID texts is expected to enhance the distinguishability between ID and OoD texts, as illustrated in Fig.~\ref{Fig1: a} (bottom).

To quantify and analyze the modality gap, we compute the uniformity and alignment measures using ImageNet-1k validation data. These measures are computed as follows~\cite{oh2024geodesic}:
\begin{equation}
\mathrm{Uniformity}:= -\log \mathbb{E}_p \left[ \exp \left( -2 \| e_i - e_j \|_2^2 \right) \right],
\label{eq uniformity}
\end{equation}
\begin{equation}
\mathrm{Alignment}:= -\mathbb{E}_p \left[ \| i_i - t_i \|_2^2 - \min_{j \neq i} \| i_i-t_j \|_2^2 \right].
\label{eq alignment}
\end{equation}
Embeddings \( e_i \) and \( e_j \) can represent either ID image embeddings \( i_i, i_j \) or ID text embeddings \( t_i \), \( t_j \), or any combination thereof. 
We explore variations in the input combinations of Eqs.~\ref{eq uniformity} and~\ref{eq alignment} to quantitatively assess different aspects of uniformity and alignment in the embedding space.

We consider four specific variants of the uniformity metric and two of the alignment metric. To assess the spread among all ID embeddings, we calculate the uniformity value (denoted as Uni-All) as in Eq.~\ref{eq uniformity}, where $e_i, e_j \in \{ i_i \}_{i=1}^n \cup \{ t_j \}_{j=1}^n$. Uni-I and Uni-T represent uniformity measured only among ID image embeddings (i.e.,  $e_i, e_j \in \{ i_i \}_{i=1}^n$) and ID text embeddings (i.e., $e_i, e_j \in \{ t_i \}_{i=1}^n$), respectively. We also compute the uniformity of cross-modal embeddings (Uni-CM) across all ID image-text pairs, where $e_i \in \{ i_i \}_{i=1}^n$ and $e_j \in \{ t_j \}_{j=1}^n$, and Uni-CMM, which uses only matching ID image-text pairs from Uni-CM, i.e., $(e_i,e_j) \in \{(i_i,t_i) \}_{i=1}^n$. Both metrics serve as indicators of the modality gap. When the Uni-CM value is comparable to Uni-All, Uni-T, and Uni-I, and the Uni-CMM value is lower than these metrics, this indicates a reduced modality gap.

The uniformity metrics quantify only the distance within the ID embedding space and do not capture the critical distance to OoD texts.
Therefore, we measure alignment metrics from two different perspectives using Eq.~\ref{eq alignment}. The first term represents the distance between the matching ID image and text embeddings. Align-ID and Align-OoD are calculated by setting $t_j$ in the second term (i.e., minimum distance) to ID and OoD text embeddings, respectively. Given an ID image, Align-ID indicates the degree to which the matching ID text embedding is distinguishable from other ID text embeddings, while Align-OoD represents the distinguishability of the matching ID text embedding from other OoD text embeddings.

\begin{table}[h!]
\vspace{-10pt}
\fontsize{8pt}{10pt}\selectfont
\setlength{\tabcolsep}{0.4mm}
  \hfill
    \centering
    \caption{Summary of uniformity and alignment values}
    \label{tab: Uniformity}
    \begin{tabular}{lccccccc} 
        \\\toprule
        & \multicolumn{5}{c}{\textbf{Uniformity}} & \multicolumn{2}{c}{\textbf{Alignment}} \\
        \cmidrule(lr){2-6} \cmidrule(lr){7-8}
        \textbf{Methods}      & Uni-All & Uni-I & Uni-T & Uni-CM & Uni-CMM & Align-ID & Align-OoD \\ 
        \midrule
        ZS & 1.594 & 0.874 & 1.320 & 2.346 & 2.141 & 0.023 & 0.035 \\  
        CoOp & 1.702 & 0.874 & 1.701 & 2.343 & 2.073 & 0.032 & 0.069 \\
        LoCoOp & 1.419 & 0.874 & 0.904 & 2.290 & 2.080 & 0.024 & 0.066 \\
        FLYP & 1.821 & 1.395 & 1.215 & 2.437 & 2.124 & 0.052 & 0.068 \\
        $m^2\text{-mix}$ &  1.208 & 0.947 & 1.441 & 1.584 & 1.014 & 0.104 & 0.130\\
        \rowcolor[gray]{0.9}CMA & 0.862 & 0.719 & 0.993 & 1.323 & 0.725 & 0.114 & 0.138\\
        \bottomrule
        \label{tab:uniformity alignment}
    \end{tabular}
    \vspace{-10pt}
\end{table}

In Table~\ref{tab: Uniformity}, we observe that CMA achieves the smallest Uni-CM and Uni-CMM values, as well as the highest Align-ID and Align-OoD values among all methods. These results demonstrate CMA's effectiveness in both OoDD performance and ID accuracy. Notably, Uni-CMM is much smaller than Uni-T and is close to Uni-I, highlighting CMA’s strength in aligning cross-modal embeddings of ID data, as illustrated in Figs.~\ref{Fig1: a} (bottom) and~\ref{fig:DOSNES}. 

In contrast, other methods show larger modality gaps. For example, Uni-CM in ZS is approximately 2.7 and 1.8 times larger than Uni-I and Uni-T, respectively. Although Uni-CMM is smaller than Uni-CM, it remains significantly larger than intra-modal distances, indicating that matching image-text pairs are still considerably more distant than embeddings within each modality. This supports the scenario depicted in Fig.~\ref{Fig1: a} (top). Moreover, FLYP shows Uni-CM and Uni-CMM values that exceed those of Uni-I and Uni-T, despite fine-tuning both encoders. This finding explains why FLYP$\mbox{\scriptsize{NegLabel}}$ performs worse than ZS NegLabel on the MOS benchmark datasets, even though it achieves higher Align-ID and Align-OoD values than ZS. 
Similar patterns are observed in PL methods like CoOp$\mbox{\scriptsize{NegLabel}}$ and LoCoOp$\mbox{\scriptsize{NegLabel}}$. This analysis shows that a larger modality gap can impair the distinguishability between ID and OoD texts, even when ID image-text pairs are more aligned.
Meanwhile, $m^2$-mix achieves performance close to CMA’s results. Although not as effective as CMA, $m^2$-mix shows improved OoDD performance and a reduction in the modality gap. These results suggest that, even though it was developed for different purposes, addressing the modality gap has a significant impact on effectively utilizing OoD text.

\section{Conclusion}
\label{sec:conclusion}
In this paper, we introduce cross-modal alignment (CMA), a novel multi-modal fine-tuning (MMFT) method that achieves state-of-the-art performance in both OoDD and ID accuracy. We establish a theoretical connection between CMA and EBM by incorporating the generative term into contrastive learning. Our experimental results show how the CMA regularizer enhances the hyperspherical structure of the embedding space, reduces the modality gap, and strengthens alignment, leading to better OoD detection and ID classification. We plan to further explore the effectiveness of using auxiliary negative labels in MMFT training.

\subsection*{Acknowledgement} This work was supported by the National Research Foundation of Korea (NRF) under Grant [RS-2024-00352184] and [RS-2024-00354675] funded by the Ministry of Science and ICT (MSIT).

{
    \small
    \bibliographystyle{ieeenat_fullname}
    \bibliography{main}
}

\clearpage
\maketitlesupplementary
\renewcommand{\thesection}{\Alph{section}}

\renewcommand\thesection{A}
\hypertarget{sup:A Detailed Experimental Settings}{}
\section{Detailed Experimental Settings}
\label{sup:first_appendix}
\label{sup:experimental settings app}
\subsection{Datasets}
\hypertarget{sup:A.1 Datasets}{}

\paragraph{MOS Benchmark Datasets.} Our experiments utilize the MOS benchmark dataset~\cite{huang2021mos}, which has been used in prior studies~\cite{ming2022delving,wang2023clipn,jiang2024negative,miyai2024locoop}. The MOS benchmark includes ImageNet-1k~\cite{deng2009imagenet} as the in-distribution (ID) dataset with a validation set with 50,000 images. The out-of-distribution (OoD) dataset consists of four datasets: SUN~\cite{xiao2010sun}, Places~\cite{zhou2017places}, Textures~\cite{cimpoi2014describing}, and iNaturalist~\cite{van2018inaturalist}, with no overlap with ImageNet-1k. The test OoD datasets include 10,000 images each from iNaturalist, SUN, Places, and 5,640 images from Textures.

\paragraph{OpenOOD v1.5 Benchmark Datasets.} OpenOOD v1.5~\cite{zhang2023openood} includes six benchmarks: four standard OoDD benchmarks and two full-spectrum OoDD benchmarks. 
For our experiments, we utilize the standard OoDD benchmark with ImageNet-1k, which consists of 45,000 images for testing and 5,000 images for validation. This benchmark dataset includes two scenarios: Near-OoD and Far-OoD scenarios. The Near-OoD scenario includes SSB-hard~\cite{vaze2021open} (49,000 images across 980 categories) and NINCO~\cite{bitterwolf2023or} (5,879 images), while the Far-OoD scenario contains iNaturalist (10,000 images), Textures (5,640 images), and OpenImage-O~\cite{wang2022vim} (1,763 images).

\subsection{Implementation details}
\label{sup:Implementation details}
\hypertarget{sup:A.2 Implementation details}{}

Our proposed method, Cross-Modal Alignment (CMA), was implemented using Python 3.9.18 and PyTorch 1.12.0+cu116. All experiments were conducted on 8 NVIDIA A6000 GPUs, each with 48GB of memory, running on Ubuntu 20.04.6 LTS.

We conduct experiments across \textit{zero-shot} (ZS), \textit{prompt learning} (PL), \textit{single-modal fine-tuning} (SMFT), and \textit{multi-modal fine-tuning} (MMFT), following the default configurations of each baseline for fair comparison. For ZS, we use the post-hoc methods MCM~\cite{ming2022delving} and NegLabel~\cite{jiang2024negative}, which leverage text information without additional training. These OoD scoring methods are also employed for PL and MMFT experiments. 

As in previous works~\cite{ming2022delving,jiang2024negative}, we use the prompts ``a photo of a [label]'' for MCM in the ZS setting and ``The nice [label]'' for NegLabel with temperature scaling set to 1 and 0.01, respectively. NegLabel~\cite{jiang2024negative} highlights that word choice in prompt engineering significantly impacts OoD detection performance: negative prompts degrade performance, positive prompts improve it, and neutral prompts require careful tuning. Among these, the ``The nice [label]'' template provides optimal results in NegLabel. However, in MMFT methods such as FLYP~\cite{10205046}, template choice does not lead to significant accuracy variations. To explore this further, we conduct an ablation study using three different prompts, finding that the prompt choice does not significantly affect OoDD performance in MMFT. Detailed results are provided in Appendix~\ref{abl:prompt change}.

Unlike MCM, NegLabel requires additional hyperparameters due to its use of negative texts. Specifically, negative labels are mined from large word corpora like WordNet~\cite{fellbaum1998wordnet}. We follow the NegMining from \citet{jiang2024negative}, which extracts $M=10,000$ negative labels with a percentile $\eta=0.05$ from WordNet. The extracted texts are then used for OoD scoring, as described in Eq.~\ref{eq:ood score}. Additionally, the results in Tables~\ref{tab:table1} and~\ref{tab:OpenOOD v1.5} do not incorporate the grouping strategy. Detailed results on the grouping strategy are provided in Appendix~\ref{abl:grouping strategy}.

For CoOp~\cite{zhou2022learning} and LoCoOp~\cite{miyai2024locoop}, we also adopt the settings proposed in the original papers. However, PL employs few-shot fine-tuning with a 16-shot setting, making direct comparisons with SMFT and MMFT potentially unfair. To address this, we not only follow the original settings but also evaluate with increasing shot counts, up to the full-shot setting where all ID data are utilized, as detailed in Appendix~\ref{abl:coop locoop}. In the 16-shot setting, we report the averaged results from three repeated experiments with seeds 0, 1, and 2 using the original codebase~\cite{zhou2022learning,miyai2024locoop}. For other shot settings, we do not perform repeated experiments.

In SMFT, we use LP, FFT, and LP-FT~\cite{kumar2021fine}, each of which exclusively utilizes the visual encoder without relying on textual information. Following~\citet{10205046}, we perform a hyperparameter sweep with learning rates \{1e-4, 1e-5, 1e-6\} and weight decay values \{0.0, 0.1\}. Models are trained for 10 epochs, selecting the best based on in-distribution (ID) accuracy, as OoD data is not directly available for validation in real-world OoDD settings. Through this procedure, the learning rates are set to 1e-4 for LP and 1e-5 for FFT and LP-FT, with a weight decay of 0.0.  

For MMFT, we compare FLYP~\cite{10205046} and $m^2$-mix~\cite{oh2024geodesic} with our proposed approach, using the same hyperparameter sweep as SMFT. Our method explores learning rates \{1e-4, 1e-5, 1e-6\}, weight decay values \{0.0, 0.1\}, and alignment strengths (i.e., $\lambda$) \{1e-1, 1e-2, 1e-3\}, with a batch size of 512. Early stopping is based on the accuracy of the ID validation set. In $m^2$-mix, the mixup weight is controlled using the $\lambda$ value instead of alignment strength.

Fig.~\ref{fig:CMA:Pytorch-like pseudo-code} presents a snippet of code to illustrate our proposed method. In summary, the CLIP image and text encoders extract corresponding embeddings, which are projected into the same dimension and undergo $L_2$ normalization, projecting them onto a shared hyperspherical space. For CLIP and FLYP, contrastive learning optimizes cosine similarity by increasing it for matching image-text pairs while reducing it for non-matching pairs. 
Our method extends this approach by calculating \texttt{CMA\_text} and \texttt{CMA\_img} to derive the \texttt{total\_loss}. When the alignment strength parameter $\lambda$ is set to 0, the training process is equivalent to FLYP.

\begin{figure}[h]
\centering
    \centering
    \lstset{
        language=Python,
        basicstyle=\scriptsize\ttfamily,
        frame=single,
        keywordstyle=\color{blue},
        commentstyle=\color{blue},
        stringstyle=\color{red},
        breaklines=true,
        breakatwhitespace=true,
        keepspaces=true,
        columns=flexible,
        linewidth=\linewidth
    }
    \begin{lstlisting}
# extract image and text embeddings
img_emb, text_emb, scale = model(images, texts)

# joint hypersphirical embeddings
img_emb /= img_emb.norm(dim=-1, keepdim=True)
text_emb /= text_emb.norm(dim=-1, keepdim=True)

# scaled cosine similarity
logits = (scale) * (img_emb @ text_emb.T)

# clip symmetric loss function
gt = torch.arange(bs)
img_loss = Cross_Entropy_Loss(logits, gt, axis=0)
text_loss = Cross_Entropy_loss(logits, gt, axis=1)

# cross-modal-alignment regularization
CMA_img = -torch.logsumexp(logits.per_img,dim=1)
CMA_text = -torch.logsumexp(logits.per_text,dim=1)

# total CMA loss
total_loss = (img_loss + args.lam * CMA_img.mean())/2 
           + (text_loss + args.lam * CMA_text.mean())/2 \end{lstlisting}
    \caption{Pytorch-like pseudo-code of CMA}
    \label{fig:CMA:Pytorch-like pseudo-code}
\end{figure}
% \clearpage

\clearpage
\renewcommand\thesection{B}
\hypertarget{sup:B Additional Ablations}{}
\section{Additional Ablations}
\label{abl:ablation}
\hypertarget{sup:B.1 Ablation study of CoOp}{}
\subsection{PL results with various-shot settings} 
\label{abl:coop locoop}

In Tables~\ref{tab:table1} and~\ref{tab:OpenOOD v1.5} of the main paper, we present PL results based on the default 16-shot settings from CoOp~\cite{zhou2022learning} and LoCoOp~\cite{miyai2024locoop}. To ensure a fair comparison, we extend these implementations to consider additional shot settings, including full-shot, and compare them with SMFT and MMFT, which employ full-shot configurations. Specifically, we conduct experiments using 256, 512, 1024, and full-shot settings, as shown in Tables~\ref{shot test 1}, \ref{shot test2}, and~\ref{tab:CoOp LoCoOp ID accuracy}. All settings use a batch size of 512, consistent with the SMFT and MMFT configurations.

Our observations reveal that increasing the number of shots generally improves both OoDD performance and ID accuracy, suggesting that prompt learning benefits from more training data. However, full-shot settings do not always yield better results; in some benchmarks, performance at full-shot is even worse than ZS. Additionally, the observed improvements are not sufficient to outperform other baselines.

For CoOp$\mbox{\scriptsize{NegLabel}}$, the highest ID accuracy of 74.07\% is achieved in the 1024-shot setting, as shown in Table~\ref{tab:CoOp LoCoOp ID accuracy}. In contrast, the best OoDD performance is observed in the 256-shot setting on the MOS benchmark and the 512-shot setting on the OpenOOD v1.5 benchmark. These results indicate that while increasing the number of shots from 16 to full-shot provides incremental gains, determining an optimal setting remains difficult. Nonetheless, our approach consistently outperforms CoOp and LoCoOp across all shot settings.
\begin{table*}[]
\centering
\caption{OoDD performance across different shot settings (16-, 256-, 512-, 1024-, and full-shot) for CoOp and LoCoOp on the MOS benchmark}
\label{shot test 1}
\fontsize{9pt}{10pt}\selectfont
\setlength{\tabcolsep}{0.4mm}
\begin{tabular}{lcccccccccc}
\specialrule{.1em}{.05em}{.05em}
          & \multicolumn{2}{c}{iNaturalist} & \multicolumn{2}{c}{SUN} & \multicolumn{2}{c}{Places} & \multicolumn{2}{c}{Textures} & \multicolumn{2}{c}{Average} \\ \cline{2-11} 
\textbf{Methods}    & \scriptsize{FPR95↓}          & \scriptsize{AUROC↑}          & \scriptsize{FPR95↓}      & \scriptsize{AUROC↑}      & \scriptsize{FPR95↓}        & \scriptsize{AUROC↑}       & \scriptsize{FPR95↓}         & \scriptsize{AUROC↑}        & \scriptsize{FPR95↓}        & \scriptsize{AUROC↑}        \\ \hline
\textit{Zero-Shot} &                &                &            &            &              &             &               &              &              &              \\

MCM & 32.28 & 94.40   & 39.33  & 92.28   & 44.94& 89.83   & 57.98&85.99  & 43.63&90.63        \\
NegLabel& \textbf{2.30} & \textbf{99.37} & \textbf{23.23} & \textbf{95.14} & 39.85 & 90.98 & 46.49 & 89.64 & 27.97 & 93.78         \\
\hline
\textit{16-shot}    &      &      &     &     &      &         &          &       &       &   \\

CoOp$\mbox{\scriptsize{MCM}}$& 26.37 &  94.49  & 35.23  & 92.59 & 43.29& 89.67  & 41.47& 90.62 & 36.59 & 91.84\\

CoOp$\mbox{\scriptsize{NegLabel}}$&  4.95 & 98.90  & \underline{25.76}  & \underline{94.59}   & \underline{30.07} & 93.33   & 44.35 & 89.59 & 26.28  & 94.10\\

LoCoOp$\mbox{\scriptsize{MCM}}$ & 23.08 & 95.46 & 33.39 & 93.25 & 40.74 & 90.52 & 40.75 & 91.14 & 34.49 & 92.59 \\

LoCoOp$\mbox{\scriptsize{NegLabel}}$& \underline{3.19} & \underline{99.25} & 46.63 & 90.58 & 55.44 & 87.65 & 46.03 & 89.85 & 37.83 & 91.83 \\

\hline
\textit{256-shot}    &      &      &     &     &      &         &          &       &       &   \\

CoOp$\mbox{\scriptsize{MCM}}$& 28.26 &94.14 &34.69 &92.83 &42.05 &90.15 &41.67 &90.53 &36.67 &91.91 \\

CoOp$\mbox{\scriptsize{NegLabel}}$& 4.28 &99.00 &29.34 &94.41 &\textbf{29.07} &\underline{94.23} &34.25 &93.01 &\textbf{24.23} &\textbf{95.16}\\

LoCoOp$\mbox{\scriptsize{MCM}}$ & 18.80 & 96.12 & 34.46 & 92.92 & 42.04 & 90.32 & 39.77 & 91.55 & 33.77 & 92.73 \\

LoCoOp$\mbox{\scriptsize{NegLabel}}$& 4.37 & 99.09 & 49.39 & 90.53 & 64.01 & 85.82 & 51.45 & 88.95 & 42.31 & 91.10 \\

\hline
\textit{512-shot}    &      &      &     &     &      &         &          &       &       &   \\

CoOp$\mbox{\scriptsize{MCM}}$& 24.78 &94.80 &33.63 &92.89 &40.61 &90.46 &39.45 &91.17 &34.62 &92.33\\
         
CoOp$\mbox{\scriptsize{NegLabel}}$& 3.59 &99.14 &34.54 &93.30 &30.80 &93.79 &\textbf{31.01} &\textbf{93.64} &\underline{24.98} &\underline{94.97}\\

LoCoOp$\mbox{\scriptsize{MCM}}$ & 22.00 & 95.50 & 30.06 & 93.80 & 36.27 & 91.37 & 40.89 & 91.38 & 32.30 & 93.02 \\

LoCoOp$\mbox{\scriptsize{NegLabel}}$& 4.85 & 98.93 & 40.15 & 92.29 & 58.99 & 86.76 & 60.78 & 85.40 & 41.19 & 90.84 \\

\hline
\textit{1024-shot}    &      &      &     &     &      &         &          &       &       &   \\

CoOp$\mbox{\scriptsize{MCM}}$& 22.83 &95.15 &33.60 &92.80 &40.96 &90.46 &39.40 &91.33 &34.20 &92.44\\

CoOp$\mbox{\scriptsize{NegLabel}}$& 4.54 &98.93 &33.76 &93.65 &30.19 &\textbf{94.26} &\underline{31.73} &\underline{93.51} &25.05 &95.09\\

LoCoOp$\mbox{\scriptsize{MCM}}$ & 22.10 & 95.27 & 32.58 & 93.58 & 38.50 & 91.16 & 39.52 & 91.52 & 33.18 & 92.88\\

LoCoOp$\mbox{\scriptsize{NegLabel}}$& 3.80 & 99.10 & 41.17 & 91.97 & 56.59 & 87.95 & 56.93 & 87.43 & 39.62 & 91.61 \\

\hline
\textit{full-shot}    &      &      &     &     &      &         &          &       &       &   \\

CoOp$\mbox{\scriptsize{MCM}}$& 23.88 &94.98 &35.74 &92.49 &41.72 &90.13 &38.93 &91.14 & 29.10& 92.19\\
         
CoOp$\mbox{\scriptsize{NegLabel}}$& 5.14 &98.87 &32.80 &93.72 &32.23 &93.80 &32.81 &93.16 & 25.75& 94.89\\

LoCoOp$\mbox{\scriptsize{MCM}}$ & 20.25 & 96.01 & 32.72 & 93.25 & 38.82 & 90.87 & 39.96 & 91.39 & 32.94 & 92.88 \\

LoCoOp$\mbox{\scriptsize{NegLabel}}$& 4.48 & 99.02 & 43.44 & 91.03 & 66.05 & 83.16 & 53.51 & 88.34 & 41.87 & 90.39\\

\specialrule{.1em}{.05em}{.05em}

\\

\end{tabular}%
\end{table*}

\begin{table*}[h!]
\centering
\caption{OoDD performance across different shot settings (16-, 256-, 512-, 1024-, and full-shot) for CoOp and LoCoOp on the OpenOOD v1.5 benchmark}
\label{shot test2}
\fontsize{9pt}{10pt}\selectfont
\setlength{\tabcolsep}{0.4mm}
\begin{tabular}{lcccccccccccccc}
\specialrule{.1em}{.05em}{.05em}
          & \multicolumn{2}{c}{SSB-hard} & \multicolumn{2}{c}{NINCO}& \multicolumn{2}{c}{Near-OoD} & \multicolumn{2}{c}{iNaturalist} & \multicolumn{2}{c}{Textures} & \multicolumn{2}{c}{Openimage-O} & \multicolumn{2}{c}{Far-OoD} \\ \cline{2-15} 
\textbf{Methods}    & \scriptsize{FPR95↓}          & \scriptsize{AUROC↑}          & \scriptsize{FPR95↓}      & \scriptsize{AUROC↑} & \scriptsize{FPR95↓}        & \scriptsize{AUROC↑}      & \scriptsize{FPR95↓}        & \scriptsize{AUROC↑}       & \scriptsize{FPR95↓}         & \scriptsize{AUROC↑}        & \scriptsize{FPR95↓}        & \scriptsize{AUROC↑}   & \scriptsize{FPR95↓}        & \scriptsize{AUROC↑}     \\ \hline
\textit{Zero-Shot} &                &                &            &            &              &             &               &              &              &              \\

MCM & 89.45 & 64.11 & 82.70 & 69.82 & 86.08 & 66.97 & 61.94 & 87.62 & 54.26 & 87.71 & 53.80 & 88.60 & 56.67 & 87.98      \\

NegLabel& 81.87 & 71.32 & 69.82 & 77.09 & 75.85 & 74.21 & \textbf{2.32} & \textbf{99.36} & 44.98 & 90.56 & 31.10 & 93.10 & 26.14 & 94.34      \\

\hline

\textit{16-shot}    &      &      &     &     &      &         &          &       &       &   \\

CoOp$\mbox{\scriptsize{MCM}}$&  86.10 &  67.72  & 77.24  &  74.59  & 81.67 &  71.16  & 26.31 & 94.50 & 38.38  & 91.92 & 37.64 & 92.08& 34.11 & 92.83 \\

CoOp$\mbox{\scriptsize{NegLabel}}$& 68.20 & 78.72  & 57.71 & 84.20  & 62.96 & 81.46 & 4.96 & 98.90 &  42.74 & 90.28 & 23.18 & 95.25 & 23.63 & 94.81 \\

LoCoOp$\mbox{\scriptsize{MCM}}$& 87.38 & 66.23 & 77.04 & 73.46 & 82.21  & 69.84 & 22.98 & 95.48 & 37.89 & 92.27 & 37.44 & 92.09 & 32.77 & 93.28 \\

LoCoOp$\mbox{\scriptsize{NegLabel}}$& 76.03 & 74.42 & 66.51 & 80.96 & 71.27 & 77.69 & 3.70 & \underline{99.16} & 43.87 & 90.59 & 29.29 & 93.65 & 25.62 & 94.47 \\
\hline
\textit{256-shot}    &      &      &     &     &      &         &          &       &       &   \\

CoOp$\mbox{\scriptsize{MCM}}$& 86.87 &68.10 &76.04 &74.52 &81.45 &71.31 &28.23 &94.16 &38.11 &91.93 &37.27 &92.22 &34.54 &92.77 \\

CoOp$\mbox{\scriptsize{NegLabel}}$& \textbf{63.67} &81.66 &52.80 &86.29 &\underline{58.23} &\underline{83.98} &4.29 &98.99 &32.51 &93.65 &20.76 &95.96 &19.19 &96.20 \\

LoCoOp$\mbox{\scriptsize{MCM}}$& 86.63 & 66.46 & 76.07 & 73.77 & 81.35 & 70.11 & 18.78 & 96.14 & 36.55 & 92.70 & 34.53 & 92.76 & 29.95 & 93.86\\

LoCoOp$\mbox{\scriptsize{NegLabel}}$& 68.88 & 81.54 & 70.86 & 79.89 & 69.87 & 80.72 & 4.40 & 99.08 & 50.31 & 89.83 & 29.87 & 93.97 & 28.19 & 94.29\\

\hline
\textit{512-shot}    &      &      &     &     &      &         &          &       &       &   \\

CoOp$\mbox{\scriptsize{MCM}}$&  85.89 &68.85 &76.06 &74.85 &80.98 &71.85 &24.75 &94.81 &35.96 &92.49 &35.28 &92.57 &32.00 &93.29 \\

CoOp$\mbox{\scriptsize{NegLabel}}$& 68.48 &79.54 &\textbf{50.01} &\textbf{87.31} &59.25 &83.43 &\underline{3.60} &99.14 &\textbf{29.28} &\textbf{94.22} &\textbf{18.90} &\underline{96.34} &\textbf{17.26} &\textbf{96.57}\\

LoCoOp$\mbox{\scriptsize{MCM}}$&86.40 & 66.30 & 75.30 & 73.94 & 80.85 & 70.12 & 21.93 & 95.51 & 37.89 & 92.55 & 34.44 & 92.77 & 31.42 & 93.61\\

LoCoOp$\mbox{\scriptsize{NegLabel}}$&  67.23 & 81.48 & 70.40 & 78.99 & 68.82 & 80.23 & 4.86 & 98.92 & 59.40 & 86.31 & 30.01 & 93.91 & 31.42 & 93.05 \\

\hline
\textit{1024-shot}    &      &      &     &     &      &         &          &       &       &   \\

CoOp$\mbox{\scriptsize{MCM}}$&  86.04 &68.76 &75.94 &75.58 &80.99 &72.17 &22.78 &95.17 &36.11 &92.56 &34.69 &92.71 &31.19 &93.48\\

CoOp$\mbox{\scriptsize{NegLabel}}$& 68.92 &79.90 & \underline{50.16} &\underline{87.23} &59.54 &83.57 &4.56 &98.93 &\underline{30.07} &\underline{94.10} &19.83 &96.20 &\underline{18.15} &\underline{96.41} \\

LoCoOp$\mbox{\scriptsize{MCM}}$& 86.37 & 66.53 & 75.78 & 74.48 & 81.07 & 70.51 & 21.94 & 95.28 & 36.55 & 92.55 & 35.59 & 92.46 & 31.36 & 93.43\\

LoCoOp$\mbox{\scriptsize{NegLabel}}$& 65.79 & \underline{82.29} & 71.85 & 77.62 & 68.82 & 79.95 & 3.84 & 99.09 & 55.83 & 88.11 & 29.45 & 93.84 & 29.71 & 93.68 \\

\hline
\textit{full-shot}    &      &      &     &     &      &         &          &       &       &   \\

CoOp$\mbox{\scriptsize{MCM}}$& 86.13 &68.92 &75.96 &75.17 &81.04 &72.04 &23.84 &94.99 &35.45 &92.41 &34.70 &92.77 &31.33 &93.39 \\

CoOp$\mbox{\scriptsize{NegLabel}}$&\underline{63.73} &\textbf{82.39} &52.27 &86.90 &\textbf{58.00} &\textbf{84.64} &5.16 &98.87 &31.10 &93.73 &\underline{19.02} &\textbf{96.42} &18.43 &96.34 \\

LoCoOp$\mbox{\scriptsize{MCM}}$& 85.89 & 67.35 & 74.84 & 75.03 & 80.36 & 71.19 & 20.17 & 96.02 & 37.09 & 92.52 & 33.62 & 92.95 & 30.29 & 93.83 \\

LoCoOp$\mbox{\scriptsize{NegLabel}}$& 69.13 & 80.92 & 72.60 & 77.84 & 70.86 & 79.38 & 4.50 & 99.01 & 51.92 & 89.25 & 30.11 & 93.68 & 28.84 & 93.98 \\

\specialrule{.1em}{.05em}{.05em}

\\

\vspace{-30pt}

\end{tabular}%
% }
\end{table*}

\begin{table*}[]
    \centering
    \caption{ID accuracy across different shot settings (16-, 256-, 512-, 1024-, and full-shot) for CoOp and LoCoOp on ImageNet-1k}
        \label{tab:CoOp LoCoOp ID accuracy}
        \begin{tabular}{lc}
        \toprule
        Methods      & Acc. \\ 
        \hline
        \textit{16-shot}  \\
        CoOp & 71.95\\
        LoCoOp & 71.72 \\
        \hline
        \textit{256-shot} \\
        CoOp & 72.96\\
        LoCoOp & 72.76 \\
        \hline
        \textit{512-shot} \\
        CoOp & 73.61\\
        LoCoOp & 73.12 \\
        \hline
        \textit{1024-shot} \\
        CoOp  & \textbf{74.07} \\
        LoCoOp  & 73.28 \\
        \hline
        \textit{full-shot} \\
        CoOp  &  \underline{73.97} \\
        LoCoOp  &  73.44\\
        \bottomrule
        \end{tabular}
\end{table*}

\hypertarget{sup:B.2 Ablation study on grouping strategy}{}
\subsection{The effect of grouping strategy}
\label{abl:grouping strategy}
The NegMining algorithm expands textual information by selecting words maximally distant from ID texts, thereby reducing the risk of high similarity between ID images and negative labels, as described in Algorithm~\ref{negmining}. However, increasing the number of negative labels raises the variance in OoD scores, which can lead to more false positives. To address this, NegLabel~\cite{jiang2024negative} has proposed a grouping strategy that divides the negative labels into multiple groups to balance the benefits of additional information with the risk of false positives. 

We report the performance of the grouping strategy proposed by NegLabel at $n=100$ in Table~\ref{tab:group stragete}. Applying the grouping strategy improves OoDD performance as shown in the table. To highlight the inherent capabilities of CMA, we do not apply additional performance-enhancing techniques, such as the grouping strategy, in our main experiments. Nevertheless, our method achieves state-of-the-art performance without the grouping strategy and shows further improvements when it is applied.

\begin{algorithm}
\caption{NegMining (proposed in NegLabel~\cite{jiang2024negative})}
\label{negmining}
\textbf{Input}: Candidate labels $\mathcal{Y}^c$, ID labels $\mathcal{Y}$, Text encoder $f^{\text{text}}$ \\
\textbf{Output}: Negative labels $\mathcal{Y}^-$
\begin{algorithmic}[1]
\STATE // Calculate text embeddings
\FOR{$y_i \in \mathcal{Y}$}
    \STATE $e_i = f^{\text{text}}(\text{prompt}(y_i))$
\ENDFOR
\FOR{$\tilde{y}_i \in \mathcal{Y}^c$}
    \STATE $\tilde{e}_i = f^{\text{text}}(\text{prompt}(\tilde{y}_i))$
    \STATE // Measure candidate-ID label distance.
    \STATE $d_i = \text{percentile}_{\eta}(\{-\cos(\tilde{e}_i, e_k)\}_{k=1}^K)$
\ENDFOR
\STATE // Choose $M$ negative labels from top-$k$ distances.
\STATE $\mathcal{Y}^- = \text{topk}([d_1, d_2, \dots, d_C], \mathcal{Y}^c, M)$
\end{algorithmic}
\end{algorithm}

\begin{table*}[]
\centering
\fontsize{9pt}{10pt}\selectfont
\setlength{\tabcolsep}{0.4mm}
\caption{The effect of the grouping strategy ($n = 100$) on the MOS benchmark. The symbol $^\star$ represents the result with the grouping strategy.}
\label{tab:group stragete}
\begin{tabular}{lcccccccccc}
\specialrule{.1em}{.05em}{.05em}
          & \multicolumn{2}{c}{iNaturalist} & \multicolumn{2}{c}{SUN} & \multicolumn{2}{c}{Places} & \multicolumn{2}{c}{Textures} & \multicolumn{2}{c}{Average} \\ \cline{2-11} 
\textbf{Methods}    & \scriptsize{FPR95↓}          & \scriptsize{AUROC↑}          & \scriptsize{FPR95↓}      & \scriptsize{AUROC↑}      & \scriptsize{FPR95↓}        & \scriptsize{AUROC↑}       & \scriptsize{FPR95↓}         & \scriptsize{AUROC↑}        & \scriptsize{FPR95↓}        & \scriptsize{AUROC↑}        \\ \hline
\textit{Zero-Shot (ZS)} &                &                &            &            &              &             &               &              &              &              \\

MCM& 32.28 & 94.40   & 39.33  & 92.28   & 44.94& 89.83   & 57.98&85.99  & 43.63&90.63        \\
NegLabel& 2.30 & 99.37 & 23.23 & 95.14 & 39.85 & 90.98 & 46.49 & 89.64 & 27.97 & 93.78         \\
NegLabel$^\star$& \underline{1.55} & 99.58 & 17.96 & 95.82 & 33.53 & 91.97 & 44.34 & 89.86 & 24.35 & 94.31       \\

\hline
\textit{Multi-modal Fine-tuning (MMFT)}    &      &      &     &     &      &         &          &       &       &   \\

FLYP$\mbox{\scriptsize{MCM}}$& 24.86 & 94.35   & 39.81  & 90.58   & 47.92 & 87.16   & 41.19 & 89.34  & 38.44  & 90.36\\
FLYP$\mbox{\scriptsize{NegLabel}}$ & 3.16 & 99.31 & 23.48 & 94.82 & 37.23 & 90.86 & 41.70 & 89.27 & 26.39 & 93.57\\
FLYP$\mbox{\scriptsize{NegLabel}}$$^\star$ & 2.41 & 99.45 & 20.38 & 95.40 & 32.64 & 91.66 & 38.49 & 89.83 & 23.48 & 94.08\\
$m^2$-mix$\mbox{\scriptsize{MCM}}$& 22.41 & 95.61 & 39.18 & 91.85 & 47.07 & 88.72 & 43.44 & 90.13 & 38.02 & 91.58\\

$m^2$-mix$\mbox{\scriptsize{NegLabel}}$& 2.39 & 99.43 & 23.03 & 94.86 & 35.55 & 91.21 & 36.65 & 90.68 & 24.40 & 94.05 \\
$m^2$-mix$\mbox{\scriptsize{NegLabel}}$$^\star$& 1.85 & 99.53 & 20.13 & 95.41 & 31.91 & 91.96 & 34.22 & 91.17 & 22.03 & 94.52 \\
\rowcolor[gray]{0.9}CMA$\mbox{\scriptsize{MCM}}$ (Ours) &22.95 & 95.65 & 40.01 & 91.78 & 48.83 & 88.41 & 44.93 & 89.87 & 39.18 & 91.43\\
\rowcolor[gray]{0.9}CMA$\mbox{\scriptsize{NegLabel}}$ (Ours)  & 1.65 & \underline{99.62}   & \underline{16.84}  & \underline{96.36}   & \underline{27.65} & \underline{93.11}   & \underline{33.58} & \underline{91.64}  & \underline{19.93}  & \underline{95.13}\\
\rowcolor[gray]{0.9}CMA$\mbox{\scriptsize{NegLabel}}$$^\star$ (Ours) &\textbf{1.38} & \textbf{99.66} & \textbf{16.11} & \textbf{96.55} & \textbf{26.52} & \textbf{93.48} & \textbf{33.09} & \textbf{91.90} & \textbf{19.27} & \textbf{95.40}\\

\specialrule{.1em}{.05em}{.05em}
\\
\vspace{-10pt}

\end{tabular}%
% }
\end{table*}

\hypertarget{sup:B.3 Ablation study on prompt change}{}
\subsection{The impact of prompts on OoDD performance}
\label{abl:prompt change}
To evaluate the impact of prompts on performance, we conduct an ablation study using three prompts: ``A photo of a [label]'', ``The nice [label]'', and no prompt, as shown in Table~\ref{tab:comparision with different prompt settings}. These prompts are derived from the ablation study of NegLabel~\cite{jiang2024negative}. Models are trained and evaluated with the same prompts. Our results indicate that altering the prompt does not lead to significant changes in performance. Specifically, in MCM, the performance difference across prompts does not exceed 1\% in terms of average AUROC and FPR95. While positive prompts demonstrate slightly better OoDD performance, the differences are not significant enough to affect its performance superiority. These results show that the choice of prompt during MMFT has a negligible impact on OoDD performance.

\begin{table*}[]
\centering
\caption{Comparison with different prompt settings (e.g., positive, neutral, and no prompts) for FLYP and CMA on the MOS benchmark}
\label{tab:comparision with different prompt settings}
% \resizebox{\textwidth}{!}{%
\begin{tabular}{lcccccccccc}
\specialrule{.1em}{.05em}{.05em}
          & \multicolumn{2}{c}{iNaturalist} & \multicolumn{2}{c}{SUN} & \multicolumn{2}{c}{Places} & \multicolumn{2}{c}{Textures} & \multicolumn{2}{c}{Average} \\ \cline{2-11} 
\textbf{Methods}    & \scriptsize{FPR95↓}          & \scriptsize{AUROC↑}          & \scriptsize{FPR95↓}      & \scriptsize{AUROC↑}      & \scriptsize{FPR95↓}        & \scriptsize{AUROC↑}       & \scriptsize{FPR95↓}         & \scriptsize{AUROC↑}        & \scriptsize{FPR95↓}        & \scriptsize{AUROC↑}        \\

\hline
\textit{``$<class>$''}    &      &      &     &     &      &         &          &       &       &   \\

FLYP$\mbox{\scriptsize{MCM}}$ & 25.42 & 94.06 & 38.93 & 90.84 & 47.29 & 87.30 & 40.60 & 89.45 & 38.06 & 90.41\\

FLYP$\mbox{\scriptsize{NegLabel}}$ &3.64 & 99.23 & 22.61 & 95.22 & 35.70 & 91.49 & 43.44 & 88.55 & 26.35 & 93.62\\

\rowcolor[gray]{0.9}CMA$\mbox{\scriptsize{MCM}}$ (Ours)  & 21.85 & 95.73 & 39.53 & 91.85 & 48.18 & 88.56 & 45.62 & 89.78 & 38.80 & 91.48 \\

\rowcolor[gray]{0.9}CMA$\mbox{\scriptsize{NegLabel}}$ (Ours)  & 2.15 & 99.55 & \underline{18.57} & \underline{96.15} & \underline{28.98} & \underline{93.05} & \underline{34.01} & \textbf{91.86} & \underline{20.93} & \textbf{95.15}\\

\hline
\textit{``a photo of a  $<class>$''}    &      &      &     &     &      &         &          &       &       &   \\

FLYP$\mbox{\scriptsize{MCM}}$ & 26.13 & 94.04 & 39.04 & 90.64 & 47.63 & 87.02 & 41.12 & 89.90 & 38.48 & 90.40\\

FLYP$\mbox{\scriptsize{NegLabel}}$ &4.51 & 99.06 & 29.23 & 93.99 & 42.58 & 89.46 & 43.83 & 88.03 & 30.04 & 92.63\\

\rowcolor[gray]{0.9}CMA$\mbox{\scriptsize{MCM}}$ (Ours)  &22.07 & 95.80 & 38.82 & 91.91 & 47.70 & 88.62 & 44.08 & 89.94 & 38.17 & 91.57\\

\rowcolor[gray]{0.9}CMA$\mbox{\scriptsize{NegLabel}}$ (Ours)  & \underline{1.92} & \underline{99.55} & 20.72 & 96.03 & 32.28 & 92.50 & 35.27 & 91.07 & 22.55 & 94.79\\

\hline
\textit{``The nice $<class>$''}    &      &      &     &     &      &         &          &       &       &   \\

FLYP$\mbox{\scriptsize{MCM}}$& 24.86 & 94.35   & 39.81  & 90.58   & 47.92 & 87.16   & 41.19 & 89.34  & 38.44  & 90.36\\

FLYP$\mbox{\scriptsize{NegLabel}}$ & 3.16 & 99.31 & 23.48 & 94.82 & 37.23 & 90.86 & 41.70 & 89.27 & 26.39 & 93.57\\

\rowcolor[gray]{0.9}CMA$\mbox{\scriptsize{MCM}}$ (Ours) &22.95 & 95.65 & 40.01 & 91.78 & 48.83 & 88.41 & 44.93 & 89.87 & 39.18 & 91.43\\

\rowcolor[gray]{0.9}CMA$\mbox{\scriptsize{NegLabel}}$ (Ours)  & \textbf{1.65} & \textbf{99.62}   & \textbf{16.84}  & \textbf{96.36}   & \textbf{27.65} & \textbf{93.11}   & \textbf{33.58} & \underline{91.64}  & \textbf{19.93}  & \underline{95.13}\\

\specialrule{.1em}{.05em}{.05em}

\\

\end{tabular}%
% }
\end{table*}

\begin{table*}[]
\centering
\setlength{\tabcolsep}{1mm}
\caption{Comparison of different $\lambda$ values for CMA on the MOS benchmark}
\label{tab: Comparison with different lambda}
\begin{tabular}{lccccccccccc}
\specialrule{.1em}{.05em}{.05em}
          & \multicolumn{2}{c}{iNaturalist} & \multicolumn{2}{c}{SUN} & \multicolumn{2}{c}{Places} & \multicolumn{2}{c}{Textures} & \multicolumn{2}{c}{Average} & \multirow{2}{*}{Acc} \\ \cline{2-11} 
\textbf{Methods}    & \scriptsize{FPR95↓}          & \scriptsize{AUROC↑}          & \scriptsize{FPR95↓}      & \scriptsize{AUROC↑}      & \scriptsize{FPR95↓}        & \scriptsize{AUROC↑}       & \scriptsize{FPR95↓}         & \scriptsize{AUROC↑}        & \scriptsize{FPR95↓}        & \scriptsize{AUROC↑}  &  \\

\hline
$\lambda = 0.1 $    &      &      &     &     &      &         &          &       &       &  \\

CMA$\mbox{\scriptsize{MCM}}$ (Ours)  & 26.20 & 94.96 & 53.54 & 86.56 & 57.73 & 83.45 & 49.73 & 88.00 & 46.80 & 88.24 & \multirow{2}{*}{81.12} \\

CMA$\mbox{\scriptsize{NegLabel}}$ (Ours)  & 5.61 & 98.91 & 32.57 & 93.75 & 40.97 & 90.44 & 47.30 & 89.68 & 31.61 & 93.19\\

\hline
$\lambda = 0.01$    &      &      &     &     &      &         &          &       &       &   \\

CMA$\mbox{\scriptsize{MCM}}$ (Ours)  &22.70 & 95.86 & 43.95 & 90.64 & 52.63 & 87.33 & 45.80 & 90.09 & 41.27 & 90.98 & \multirow{2}{*}{81.96} \\

CMA$\mbox{\scriptsize{NegLabel}}$ (Ours)  &2.89 & 99.41 & 20.24 & 95.81 & 31.11 & \underline{92.78} & 39.15 & 91.30 & 23.35 & 94.83\\

\hline
$\lambda = 0.001$    &      &      &     &     &      &         &          &       &       &   \\

CMA$\mbox{\scriptsize{MCM}}$ (Ours) &22.95 & 95.65 & 40.01 & 91.78 & 48.83 & 88.41 & 44.93 & 89.87 & 39.18 & 91.43 & \multirow{2}{*}{\textbf{82.64}}\\

CMA$\mbox{\scriptsize{NegLabel}}$ (Ours)  & \textbf{1.65} & \textbf{99.62}   & \textbf{16.84}  & \textbf{96.36}   & \textbf{27.65} & \textbf{93.11}   & \underline{33.58} & \underline{91.64}  & \textbf{19.93}  & \textbf{95.13}\\

\hline
$\lambda = 0.0005$    &      &      &     &     &      &         &          &       &       &   \\

CMA$\mbox{\scriptsize{MCM}}$ (Ours) &22.20 & 95.73 & 38.72 & 91.96 & 47.59 & 88.17 & 42.13 & 90.16 & 37.66 & 91.50 & \multirow{2}{*}{\underline{82.56}} \\

CMA$\mbox{\scriptsize{NegLabel}}$ (Ours)  & \underline{1.85} & \underline{99.61} & \underline{17.70} & \underline{96.08} & \underline{29.50} & 92.50 & \textbf{31.54} & \textbf{92.12} & \underline{20.15} & \underline{95.08}\\

\specialrule{.1em}{.05em}{.05em}

\\
\vspace{-10pt}

\end{tabular}%
% }
\end{table*}

\hypertarget{sup:B.4 Ablation study on lambda}{}
\subsection{The impact of $\lambda$ values on OoDD performance}
\label{sup:lambda}
We select the $\lambda$ value based on ID accuracy, as actual OoD data is not available for evaluation. To determine the optimal value of $\lambda$, we compare different $\lambda$ values \{1e-1, 1e-2, 1e-3, 5e-4\}, and 1e-3 yields the highest ID accuracy, which is also aligned with the best OoDD performance, as shown in Table~\ref{tab: Comparison with different lambda}. Notably, an increase in alignment strength does not consistently improve ID accuracy or OoDD performance, highlighting the need for careful tuning. In our experiments, CMA demonstrates strong OoDD performance when the $\lambda$ value is optimized for ID accuracy, even without access to OoD data.
\clearpage

\clearpage
\hypertarget{sup:B.5 Ablation study on NearOoD scenario}{}
\subsection{Additional Near-OoD experiments}
\label{sup:Imagenet-split}

To thoroughly evaluate performance in Near-OoD scenarios, we conduct experiments on challenging ImageNet-1k splits from~\cite{palechor2023large}. Specifically, we adopt the $P_1$, $P_2$, and $P_3$ protocols in~\cite{palechor2023large}. Each of these includes known, negative, and unknown classes. For example, in $P_1$, the known classes consist of 116 fine-grained dog breeds from ImageNet. The unknown classes include 166 non-animal categories that are semantically distant from the known classes. Additionally, 67 four-legged animal classes are designated as negative classes, which are semantically closer to the known classes but remain distinct. The negative classes are originally intended to aid the model in distinguishing known classes from unknown classes during training. 

For our Near-OoD experiments, we treat negative classes, along with unknown classes, as OoD since a simple zero-shot (ZS) method using NegLabel yields near-perfect performance on $P_1$ and $P_2$, making it difficult to evaluate the benefits of MMFT approaches. 
As shown in Table~\ref{tab:table12}, ZS NegLabel achieves AUROC scores of 99.96\% and 99.42\% for $P_1$ and $P_2$, respectively. These results indicate that the unknown classes can be effectively distinguished using pre-trained textual information. Since the model already separates the unknown set too well, it becomes challenging to evaluate the contribution of textual information in MMFT. To address this, we construct more challenging splits $P'_1$, $P'_2$, and $P'_3$ by designating additional negative datasets as OoD (i.e., negative classes + unknown classes), as described in Table~\ref{tab:table11}. We perform a hyperparameter search based on FPR95 using the validation sets of known and negative classes. Note that negative classes are used only as validation/test datasets, and are not included in training.

As shown in Table~\ref{tab:table13}, our method achieves the highest AUROC scores, maintaining robust OoDD performance even under challenging conditions, while also achieving the highest accuracy among all compared methods. However, we observe that although AUROC remains higher than that of $\text{FFT}_{\text{Energy}}$ (i.e., SMFT), which does not utilize textual information, the average FPR95 is comparable. To gain a deeper understanding, we analyze each protocol in sequence.

Starting with $P_1^\prime$, we observe that $\text{FFT}_{\text{Energy}}$ underperforms in both FPR95 and AUROC compared to methods that utilize textual information through NegLabel. This can be attributed to the fact that in $P_1^\prime$, the semantic distance between unknown/negative classes and known classes is sufficiently large, allowing textual information such as negative concept labels to effectively distinguish them. This observation aligns with prior findings on the effectiveness of textual information in Far-OoD scenario.
Next, in $P_2^\prime$, we observe that all NegLabel-based methods, except for our $\text{CMA}_{\text{NegLabel}}$, underperform compared to $\text{FFT}_{\text{Energy}}$. This indicates that CMA effectively reduces the modality gap, thereby improving the utilization of textual information. Similarly, in $P_3^\prime$, while our method performs worse than SMFT in terms of FPR95, it achieves a higher AUROC score.

These findings indicate that $\text{FFT}_{\text{Energy}}$, which relies solely on visual features, can effectively distinguish between subclasses within a broader category (e.g., various types of ``Hunting Dog'' in $P_2^\prime$) solely based on visual cues. In contrast, existing NegLabel-based approaches struggle to separate ID and OoD classes when they belong to the same or semantically related categories, likely due to the modality gap. Our method addresses this challenge by mitigating the modality gap, thereby improving detection performance in Near-OoD scenarios.

\begin{table}[h]
    \centering
    \fontsize{9pt}{10pt}\selectfont
    \setlength{\tabcolsep}{1mm}
    \renewcommand{\arraystretch}{1.2} 
    \begin{tabular}{l|c|c}
        \toprule
        & \textbf{ID (Known)} & \textbf{OOD (Negative+Unknown)} \\
        \midrule
        & \multirow{2}{*}{All dog classes} & \multicolumn{1}{c}{Other 4-legged animal classes} \\
        \textbf{$P'_1$}  & \multirow{2}{*}{29055 / 5800} & \multicolumn{1}{c}{Non-animal classes}\\
        &  & 17420 / 11650 (3350+8300) \\
        \midrule
         & \multirow{2}{*}{Half of hunting dog classes} & \multicolumn{1}{c}{Half of hunting dog classes} \\
        \textbf{$P'_2$} & \multirow{2}{*}{7224 / 1500} & \multicolumn{1}{c}{Other 4-legged animal classes}\\
        &  & 7949 / 4300 (1550+2750) \\
        \midrule
        \textbf{$P'_3$} & \multicolumn{1}{c|}{Mix of common classes} & \multicolumn{1}{c}{Mix of common classes} \\
        & 38633 / 7550 & 24549 / 13050 (4850+8200) \\
        \bottomrule
    \end{tabular}
    \caption{
    More challenging ImageNet-1k splits. The numbers represent the number of validation/test samples.}
    \label{tab:table11}
\end{table}

\begin{table*}[t]
\fontsize{9pt}{10pt}\selectfont
\setlength{\tabcolsep}{0.4mm}
\centering
\caption{ZS OoDD Performance on splits $P_1$, $P_2$, and $P_3$ (ID = Known, OoD = Unknown)}
\label{tab:table12}
% \resizebox{\textwidth}{!}{%
\small{
\begin{tabular}{lcccccccc}
\toprule
& \multicolumn{2}{c}{$P_1$} & \multicolumn{2}{c}{$P_2$} & \multicolumn{2}{c}{$P_3$} & \multicolumn{2}{c}{Average} \\ \cline{2-9} 
\textbf{Methods}    & \scriptsize{FPR95$\downarrow$} & \scriptsize{AUROC$\uparrow$} & \scriptsize{FPR95$\downarrow$} & \scriptsize{AUROC$\uparrow$} & \scriptsize{FPR95$\downarrow$} & \scriptsize{AUROC$\uparrow$} & \scriptsize{FPR95$\downarrow$} & \scriptsize{AUROC$\uparrow$} 
\\ 
\midrule
\textit{Zero-Shot (ZS)} & & & & & & & & \\
MCM & 14.27 & 96.96 & 56.07 & 88.89 & 35.11 & 89.64 & 35.15  & 91.83  \\
NegLabel & \textbf{0.23} & \textbf{99.96} & \textbf{3.35} & \textbf{99.42} & \textbf{29.41} & \textbf{90.10} & \textbf{11.00} & \textbf{96.49} \\

\bottomrule

\end{tabular}%
% }
}
\vspace{-10pt}
\end{table*}

\begin{table*}[t]
\fontsize{9pt}{10pt}\selectfont
\setlength{\tabcolsep}{0.4mm}
\centering
\caption{Comparison of OoDD performance on our splits $P'_1$, $P'_2$, and $P'_3$ (ID = Known, OoD = Negative + Unknown)}
\label{tab:table13}
% \resizebox{\textwidth}{!}{%
\small{
\begin{tabular}{lcccccccccc}
\toprule
& \multicolumn{2}{c}{$P'_1$} & \multicolumn{2}{c}{$P'_2$} & \multicolumn{2}{c}{$P'_3$} & \multicolumn{2}{c}{Average} & \multirow{2}{*}{Acc}\\ \cline{2-9} 
\textbf{Methods}    & \scriptsize{FPR95$\downarrow$} & \scriptsize{AUROC$\uparrow$} & \scriptsize{FPR95$\downarrow$} & \scriptsize{AUROC$\uparrow$} & \scriptsize{FPR95$\downarrow$} & \scriptsize{AUROC$\uparrow$} & \scriptsize{FPR95$\downarrow$} & \scriptsize{AUROC$\uparrow$} 
\\ 
\midrule
\textit{Zero-Shot (ZS)} & & & & & & & & \\
MCM & 29.18 & 93.07 & 64.30 & 85.34 & 52.41 & 83.32 & 48.63  & 87.24 & \multirow{2}{*}{73.43}  \\
NegLabel & \underline{1.17} & \underline{99.74} & 24.09 & 93.34 & 48.50 & 82.90 & 24.59 & 91.99 \\
\midrule
\textit{Prompt-Learning}    &      &      &     &     &      &         &          &       \\

CoOp$\mbox{\scriptsize{MCM}}$& 24.83 & 93.91 & 63.03 & 84.75 & 52.78 & 83.64 & 46.88 & 87.43 & \multirow{2}{*}{75.10} \\

CoOp$\mbox{\scriptsize{NegLabel}}$& 1.20 & 99.73 & 26.75 & 91.65 & 48.82 & 83.72 & 25.59 & 91.70 \\

LoCoOp$\mbox{\scriptsize{MCM}}$ & 22.47 & 94.87 & 62.95 & 84.70 & 52.25 & 83.20 & 45.89 & 87.59 & \multirow{2}{*}{74.95} \\

LoCoOp$\mbox{\scriptsize{NegLabel}}$& \textbf{0.90} & \textbf{99.78} & 25.25 & 91.85 & 60.13 & 79.47 & 28.76 & 90.37\\

\midrule
\textit{Single-modal Fine-tuning (SMFT)} & & & & & & & & \\
FFT$\mbox{\scriptsize{MSP}}$ & 19.18 & 95.45 & 61.44 & 87.04 & 56.46 & 93.78 & 45.69 & 89.98 \\
FFT$\mbox{\scriptsize{ODIN}}$ & 3.24 & 99.30 & 25.88 & 93.78 & \textbf{38.00} & \underline{89.55} & 22.37 & \underline{94.21} & 84.74\\
FFT$\mbox{\scriptsize{Energy}}$ & 2.60 & 99.41 & \underline{18.19} & \underline{94.52} & \underline{39.92} & 88.40 & \textbf{20.24} & 94.11  \\
\midrule
\textit{Multi-modal Fine-tuning (MMFT)} & & & & & & & & \\
FLYP$\mbox{\scriptsize{MCM}}$ & 9.14 & 98.16 & 42.67 & 89.94 & 42.05 & 87.11 & 31.29 & 91.74 &  \multirow{2}{*}{\underline{85.30}} \\
FLYP$\mbox{\scriptsize{NegLabel}}$ & 2.33 & 99.42 & 19.35 & 94.45 & 41.76 & 86.79 & 21.15 & 93.56  \\
\cellcolor[gray]{0.9}CMA$\mbox{\scriptsize{MCM}}$ & \cellcolor[gray]{0.9}9.21 &\cellcolor[gray]{0.9}98.16 &\cellcolor[gray]{0.9}43.02 &\cellcolor[gray]{0.9}89.95 &\cellcolor[gray]{0.9}41.26 &\cellcolor[gray]{0.9}89.43 &\cellcolor[gray]{0.9}31.16 &\cellcolor[gray]{0.9}92.51 &\multirow{2}{*}{\textbf{85.55}}\\
\cellcolor[gray]{0.9}CMA$\mbox{\scriptsize{NegLabel}}$&\cellcolor[gray]{0.9}2.29 &\cellcolor[gray]{0.9}99.42 &\cellcolor[gray]{0.9}\textbf{18.07} &\cellcolor[gray]{0.9}\textbf{94.76} &\cellcolor[gray]{0.9}40.97 &\cellcolor[gray]{0.9}\textbf{89.71} &\cellcolor[gray]{0.9}\underline{20.44} &\cellcolor[gray]{0.9}\textbf{94.63} \\
\bottomrule

\end{tabular}%
% }
}
\vspace{-10pt}
\end{table*}

\renewcommand\thesection{C}
\hypertarget{sup:E Additional Visualization}{}
\section{Additional Visualization}
\label{sup:vis}
\begin{figure*}[htp]
    \centering
    \begin{subfigure}{.24\linewidth}
        \centering
        \includegraphics[width=\linewidth]{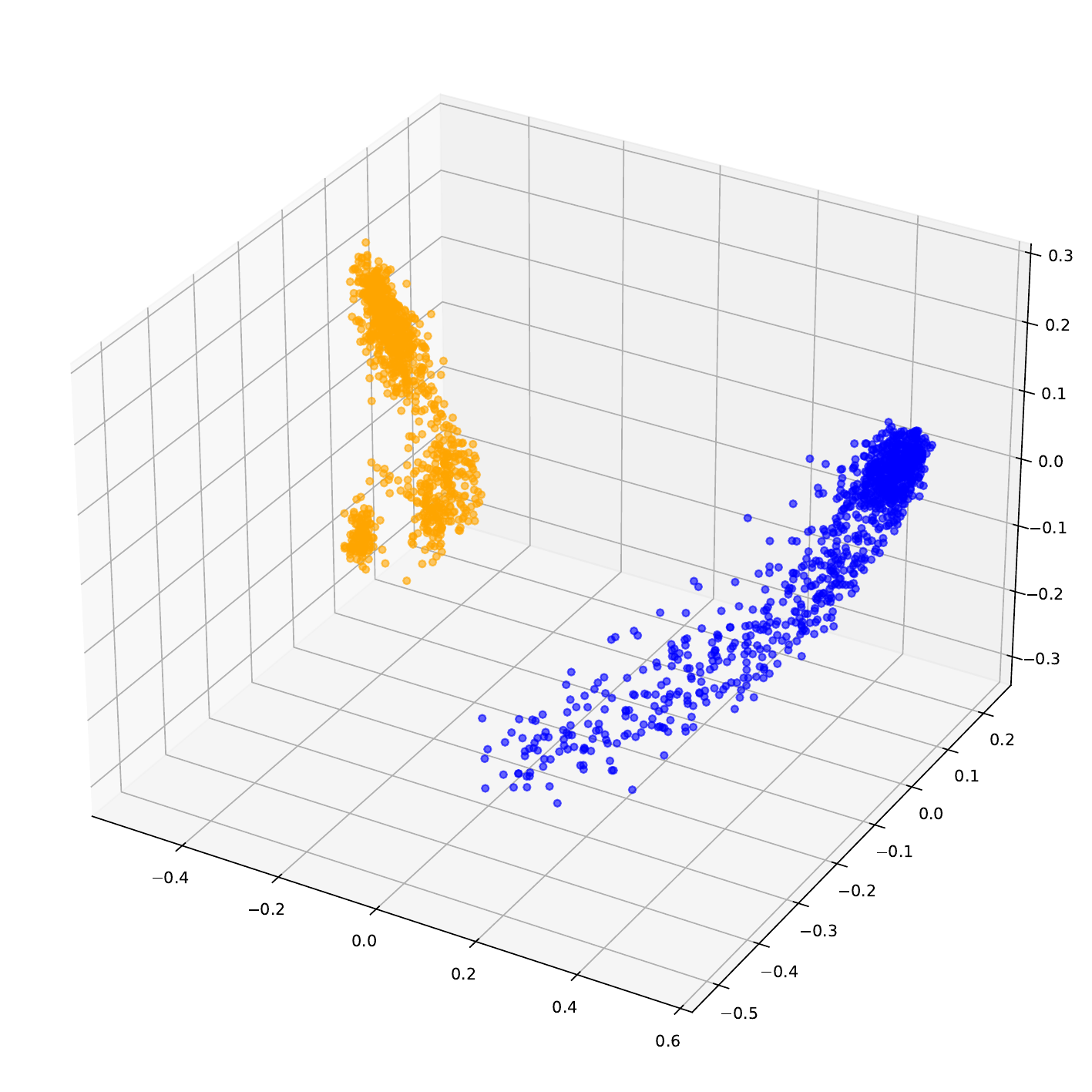}
        \caption{ZS}
        \label{fig:PCA a}
    \end{subfigure}%
    \begin{subfigure}{.24\linewidth}
        \centering
        \includegraphics[width=\linewidth]{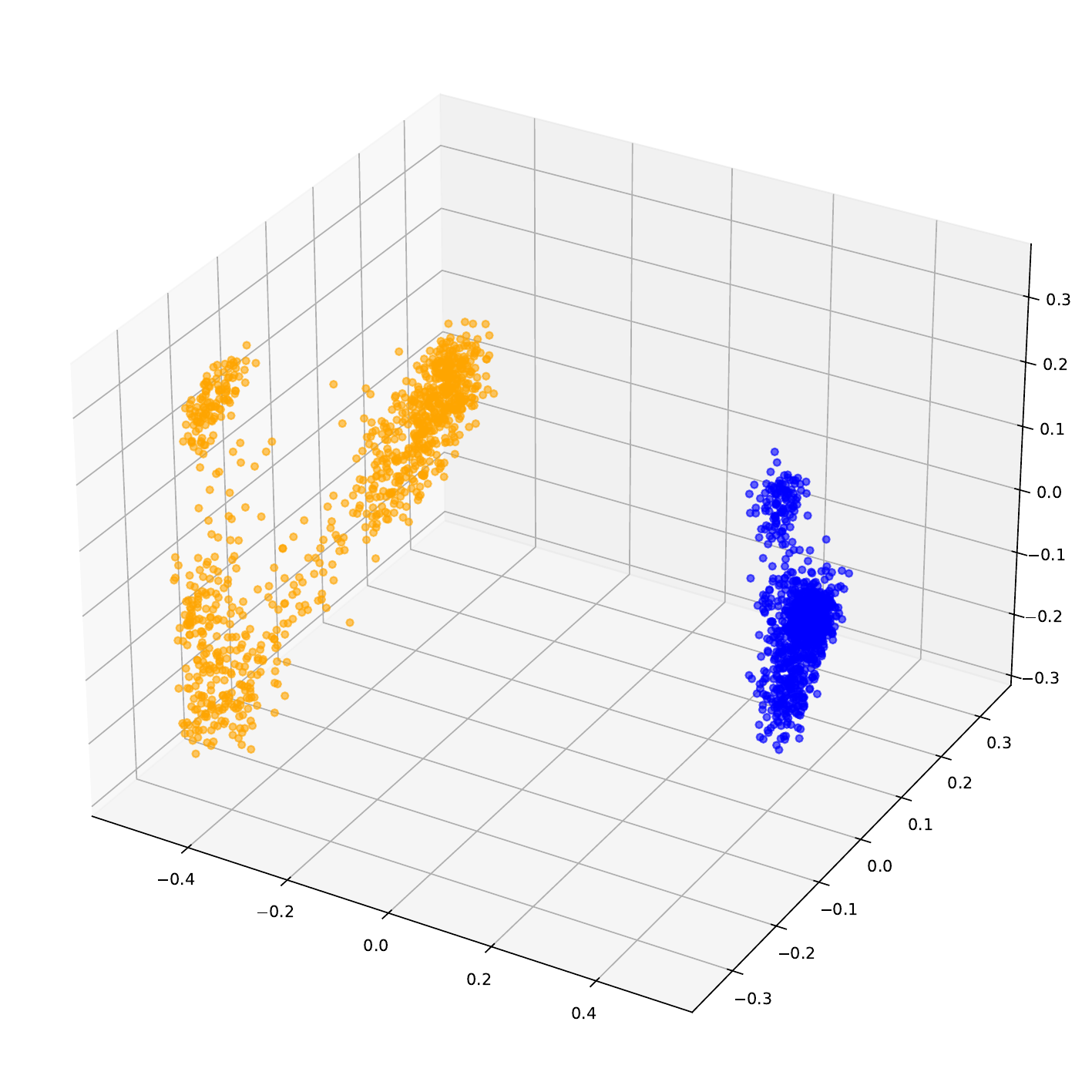}
        \caption{FLYP}
        \label{fig:PCA b}
    \end{subfigure}
    \begin{subfigure}{.24\linewidth}
        \centering
        \includegraphics[width=\linewidth]{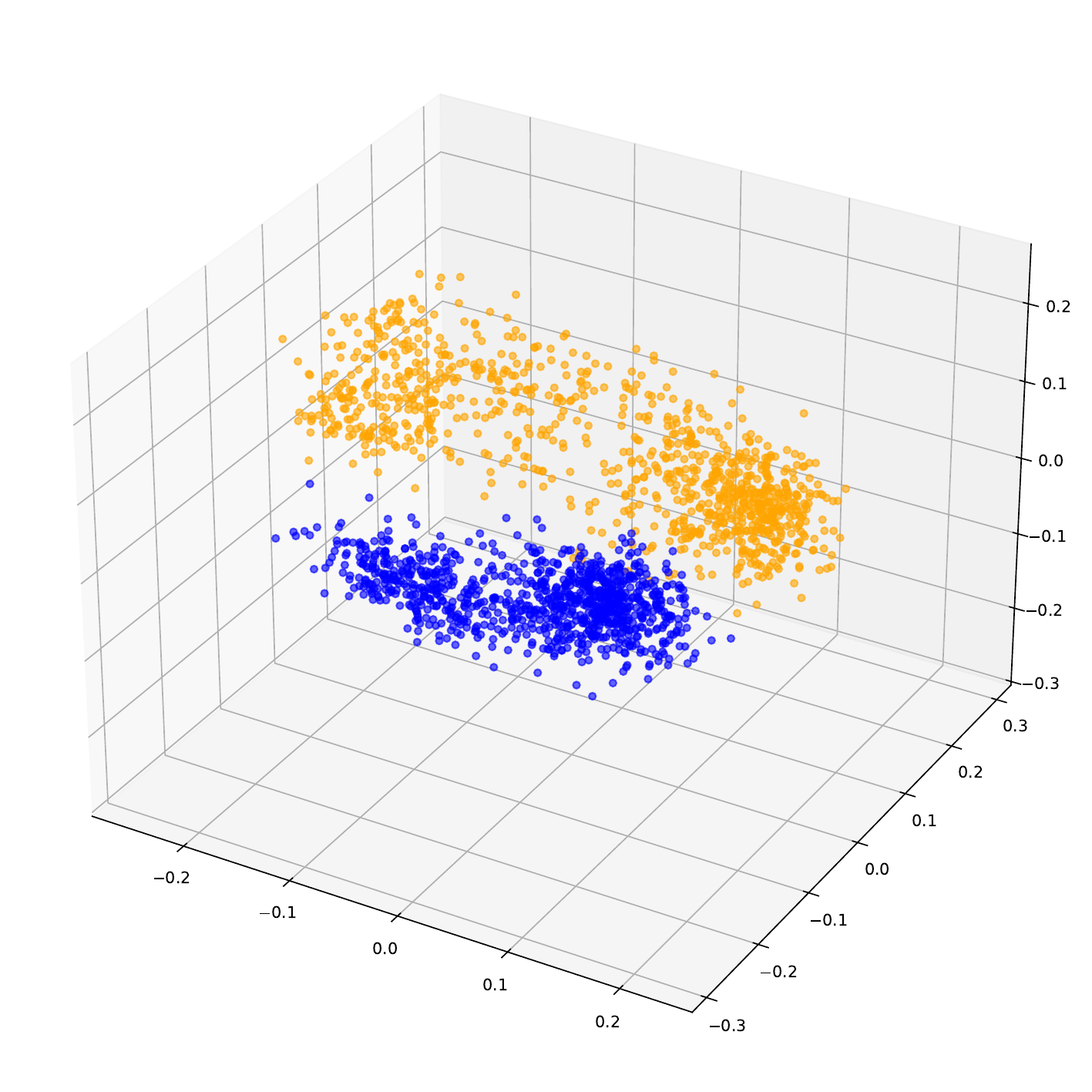}
        \caption{$m^2\text{-mix}$}
        \label{fig:PCA c}
    \end{subfigure}
    \begin{subfigure}{.24\linewidth}
        \centering
        \includegraphics[width=\linewidth]{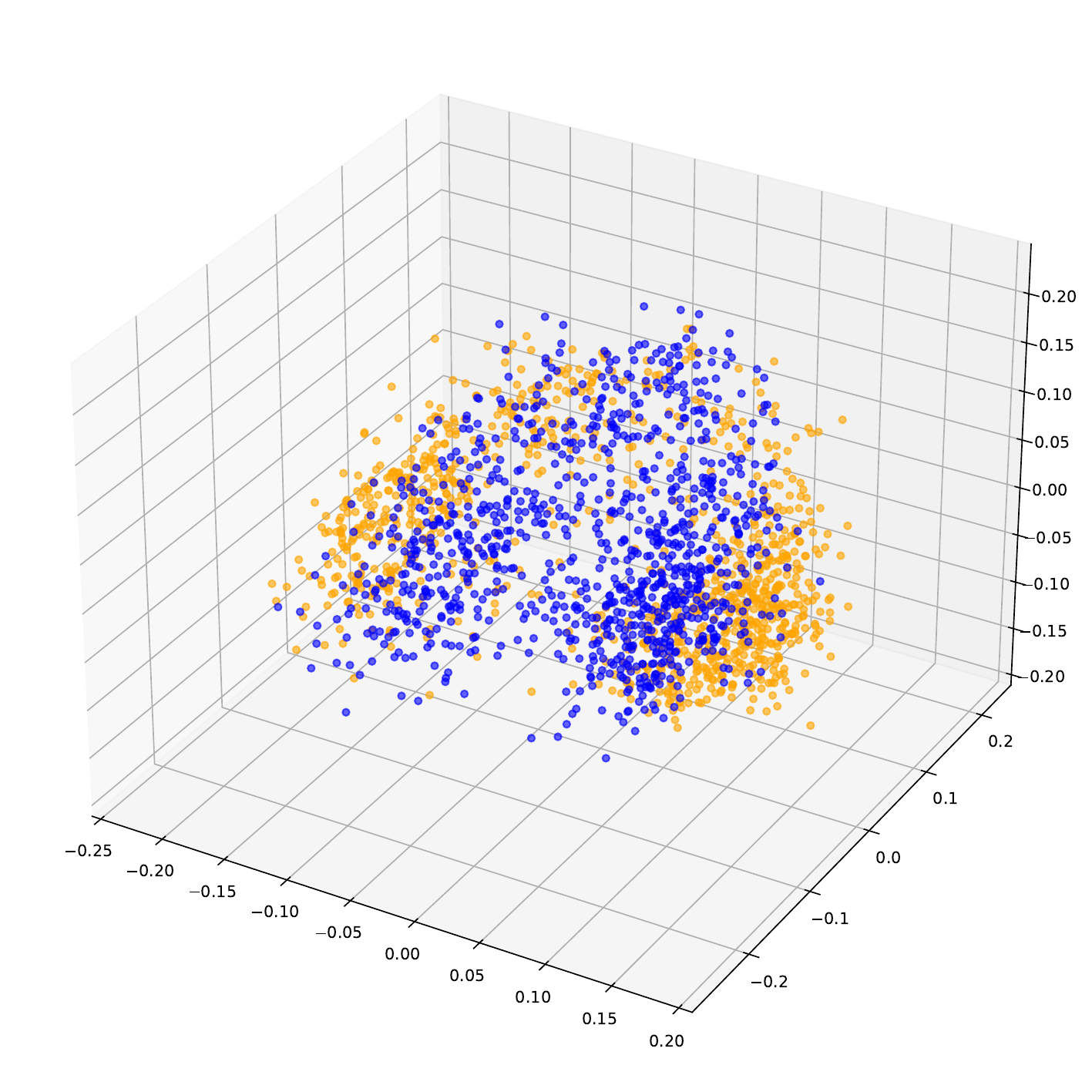}
        \caption{CMA}
        \label{fig:PCA d}
    \end{subfigure}
    \hspace{-2mm} 

    \caption{Visualization of image and text embeddings using PCA on ImageNet-1k. Orange and blue points represent ID image and ID text embeddings, respectively. }
    \label{fig:PCA}
\end{figure*}

\begin{figure*}[htp]
    \centering
    \begin{subfigure}{.24\linewidth}
        \centering
        \includegraphics[width=\linewidth]{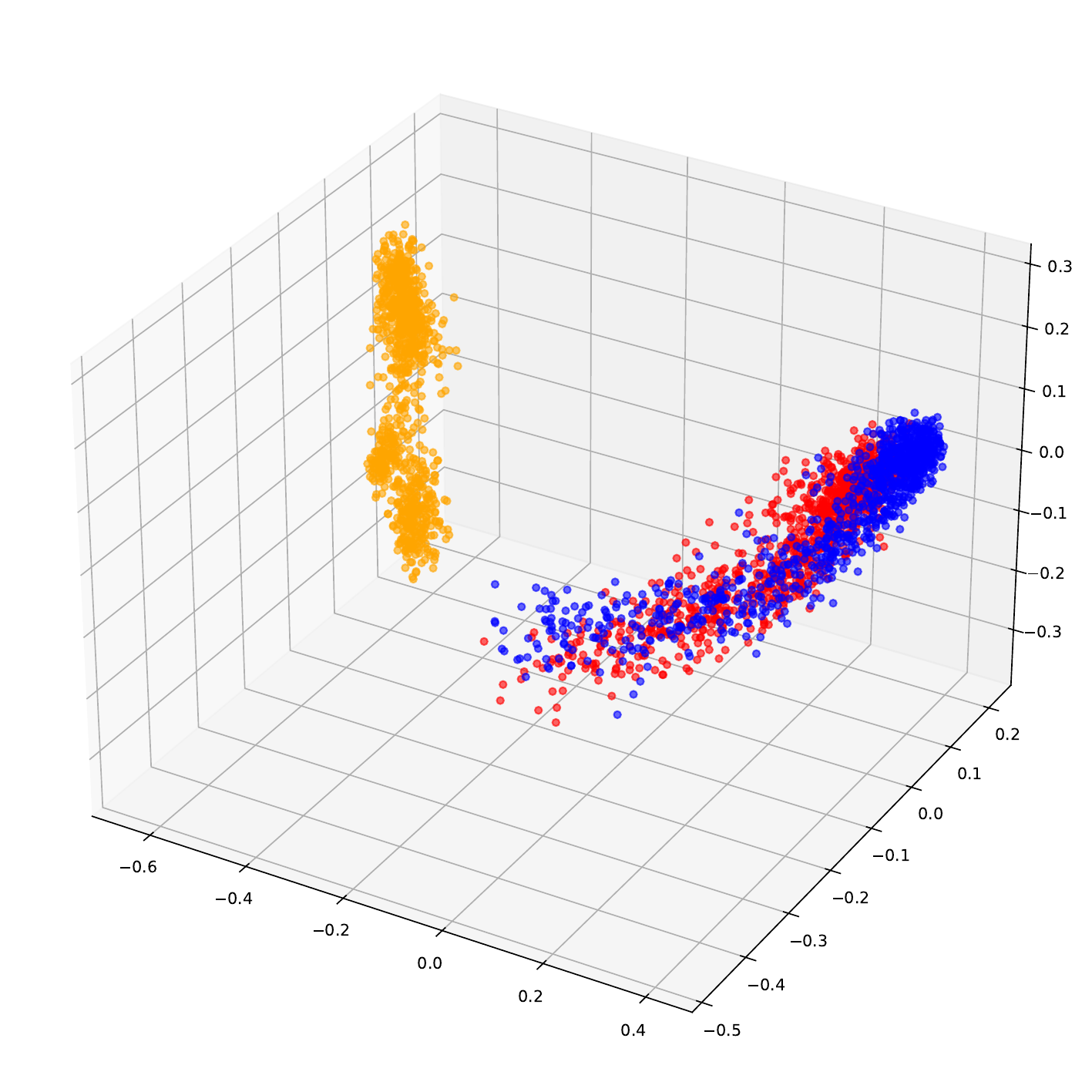}
        \caption{Zero-Shot}
        \label{fig:PCA-neg a}
    \end{subfigure}%
    \begin{subfigure}{.24\linewidth}
        \centering
        \includegraphics[width=\linewidth]{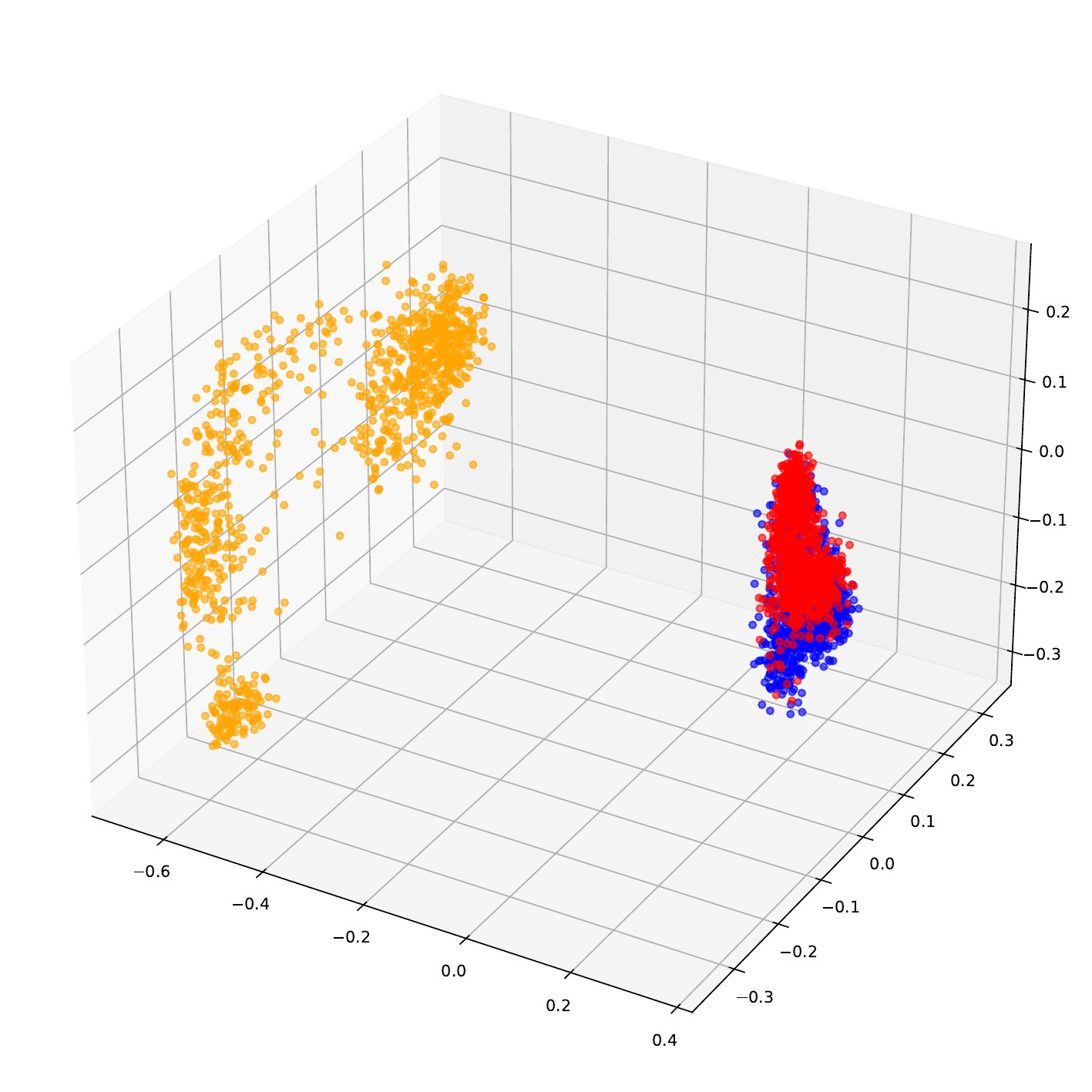}
        \caption{FLYP}
        \label{fig:PCA-neg b}
    \end{subfigure}
    \begin{subfigure}{.24\linewidth}
        \centering
        \includegraphics[width=\linewidth]{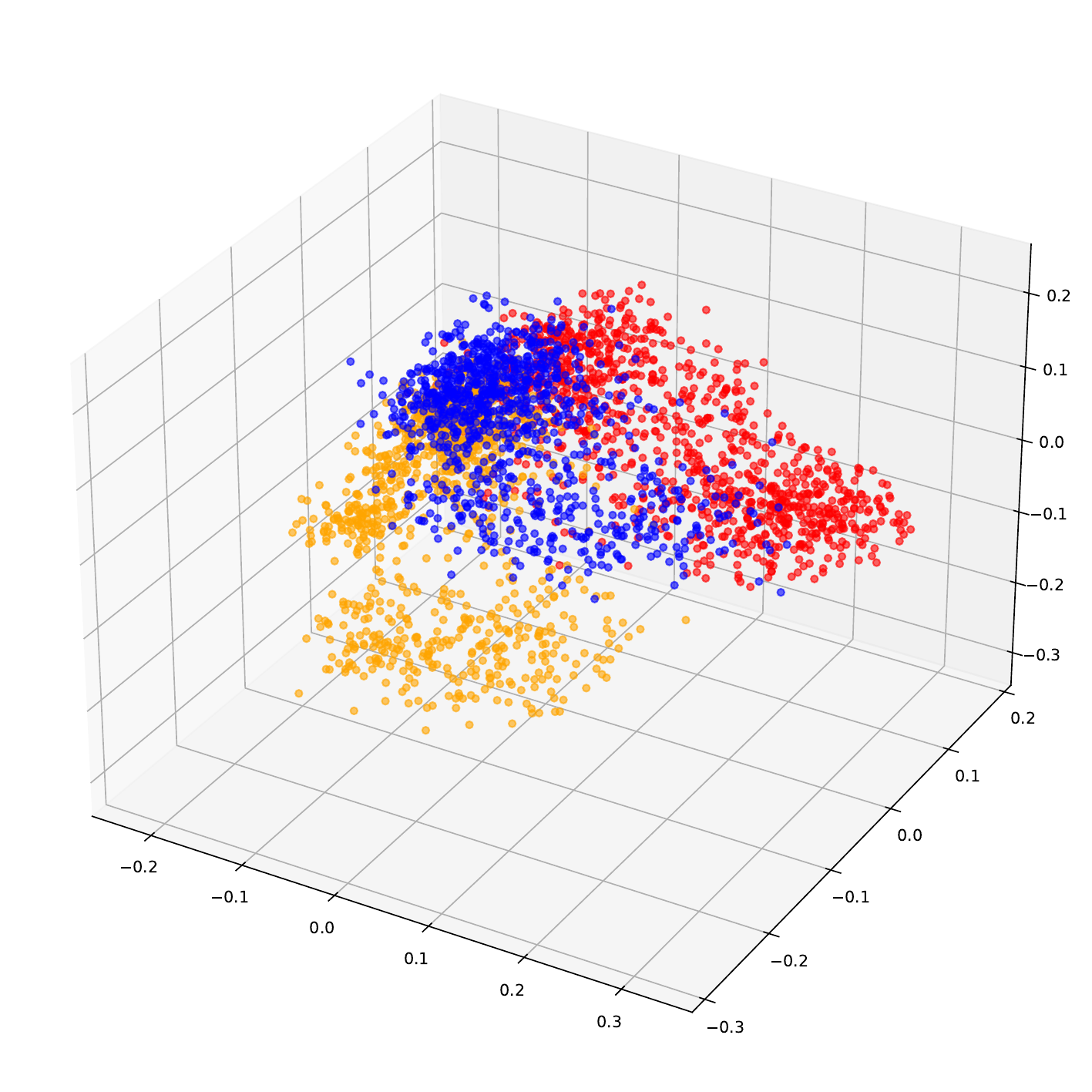}
        \caption{$m^2$-mix}
        \label{fig:PCA-neg c}
    \end{subfigure}
    \begin{subfigure}{.24\linewidth}
        \centering
        \includegraphics[width=\linewidth]{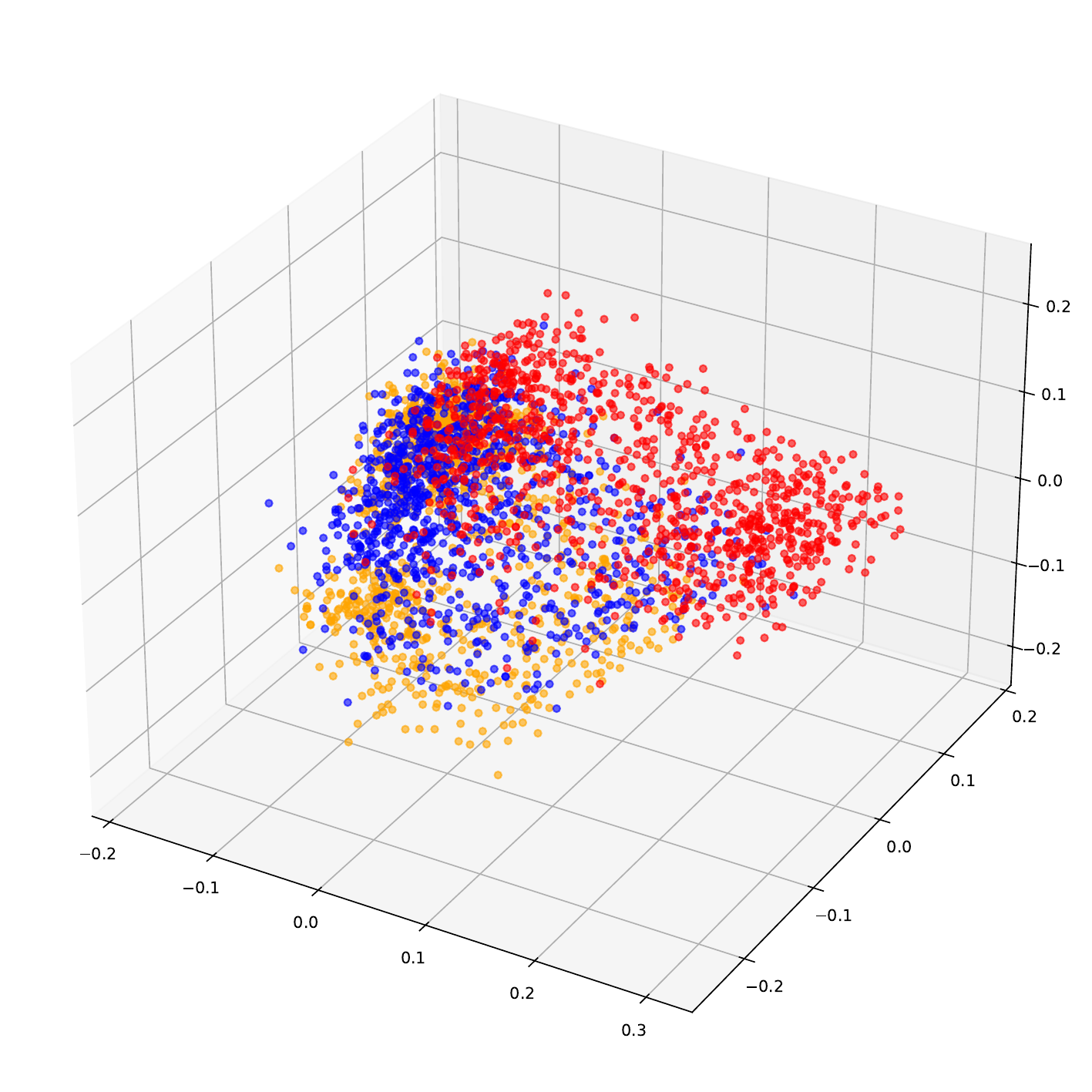}
        \caption{CMA}
        \label{fig:PCA-neg d}
    \end{subfigure}
    \hspace{-2mm} 

    \caption{Visualization of image and text embeddings using PCA on ImageNet-1k and negative texts. Orange and blue points represent ID image and ID text embeddings, respectively, while red points denote negative text embeddings.
    }\label{fig:PCA-neg}
\end{figure*}

\begin{figure*}[htp]
    \centering
    \begin{subfigure}{.24\linewidth}
        \centering
        \includegraphics[width=\linewidth]{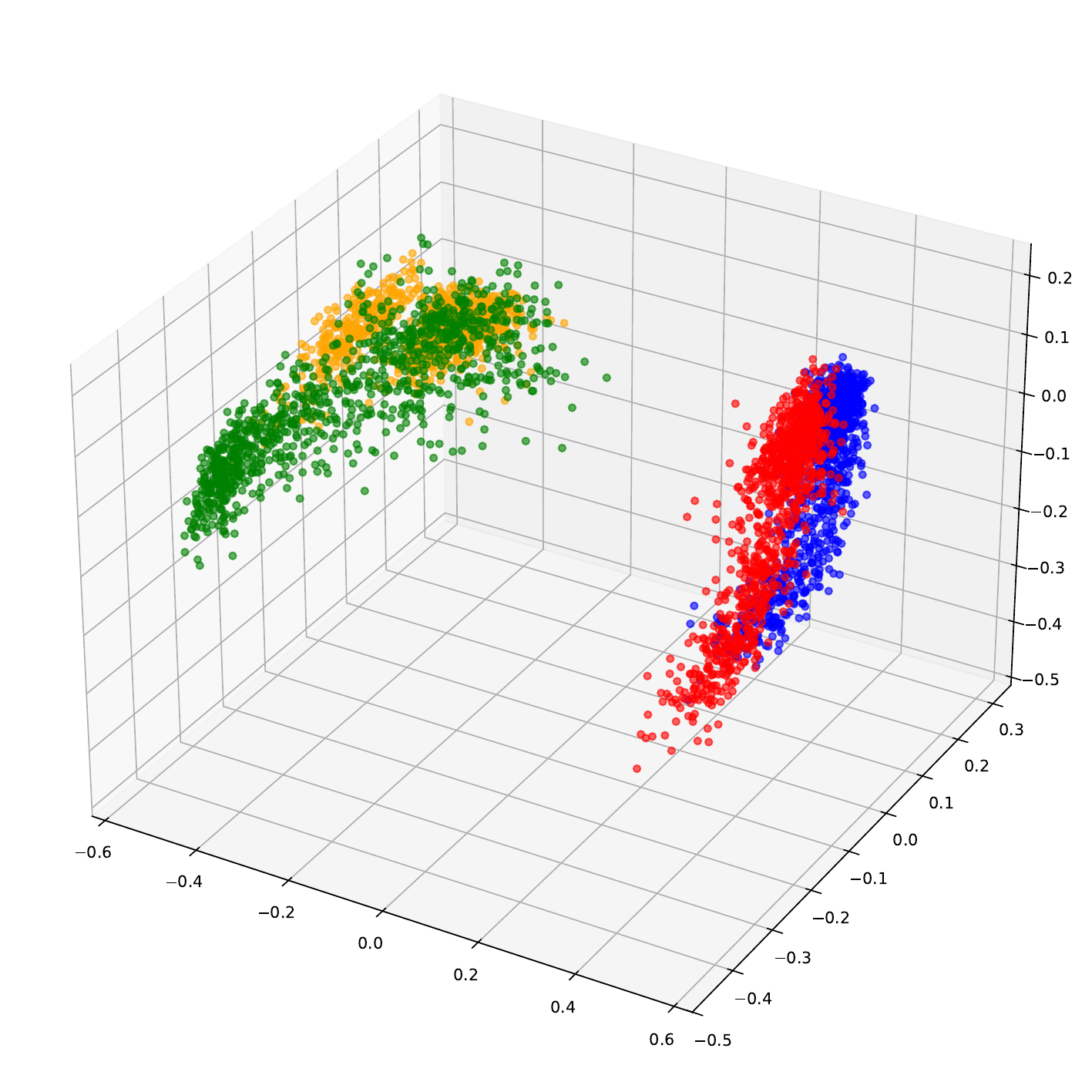}
        \caption{Zero-Shot}
        \label{fig:PCA-neg-neg a}
    \end{subfigure}%
    \begin{subfigure}{.24\linewidth}
        \centering
        \includegraphics[width=\linewidth]{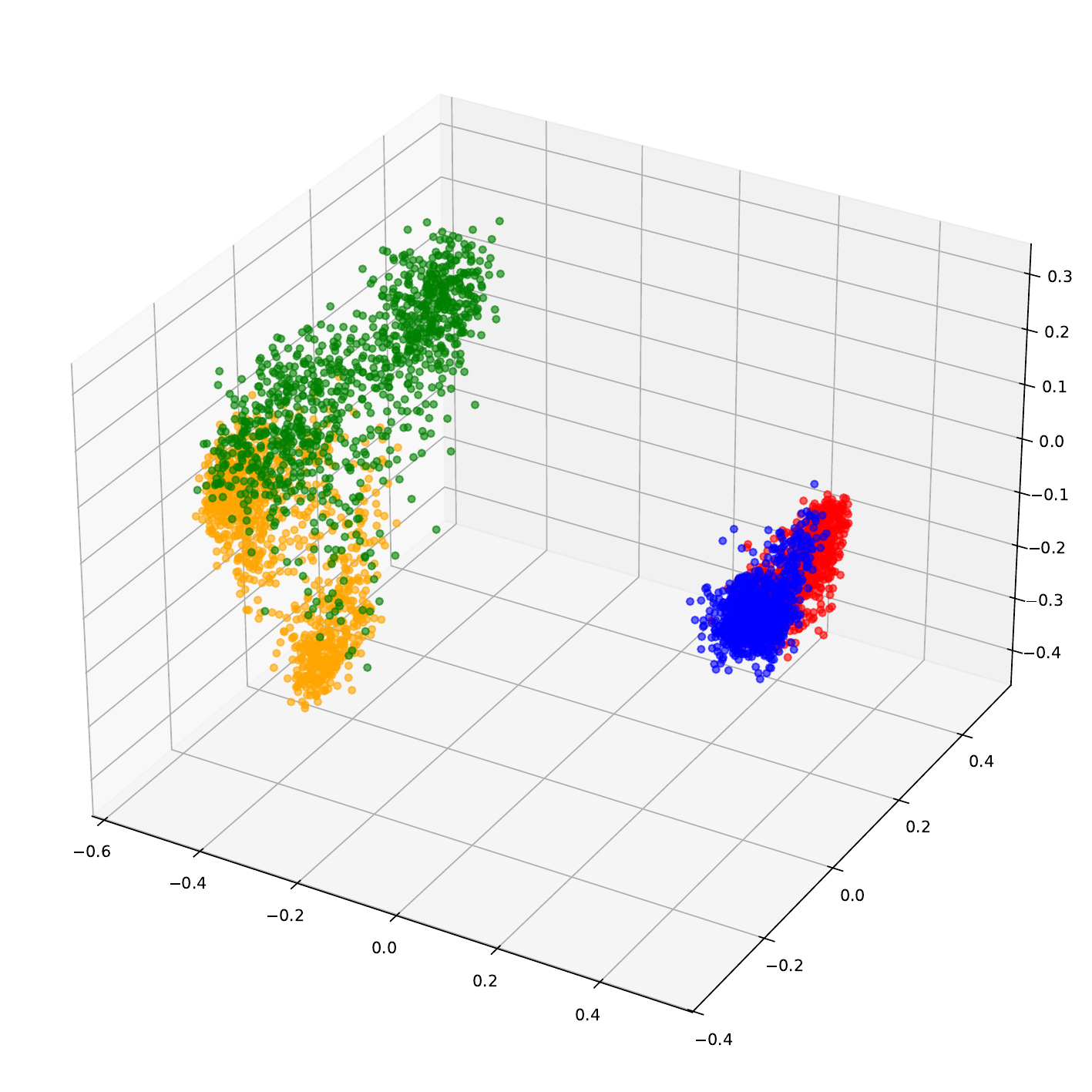}
        \caption{FLYP}
        \label{fig:PCA-neg-neg b}
    \end{subfigure}
    \begin{subfigure}{.24\linewidth}
        \centering
        \includegraphics[width=\linewidth]{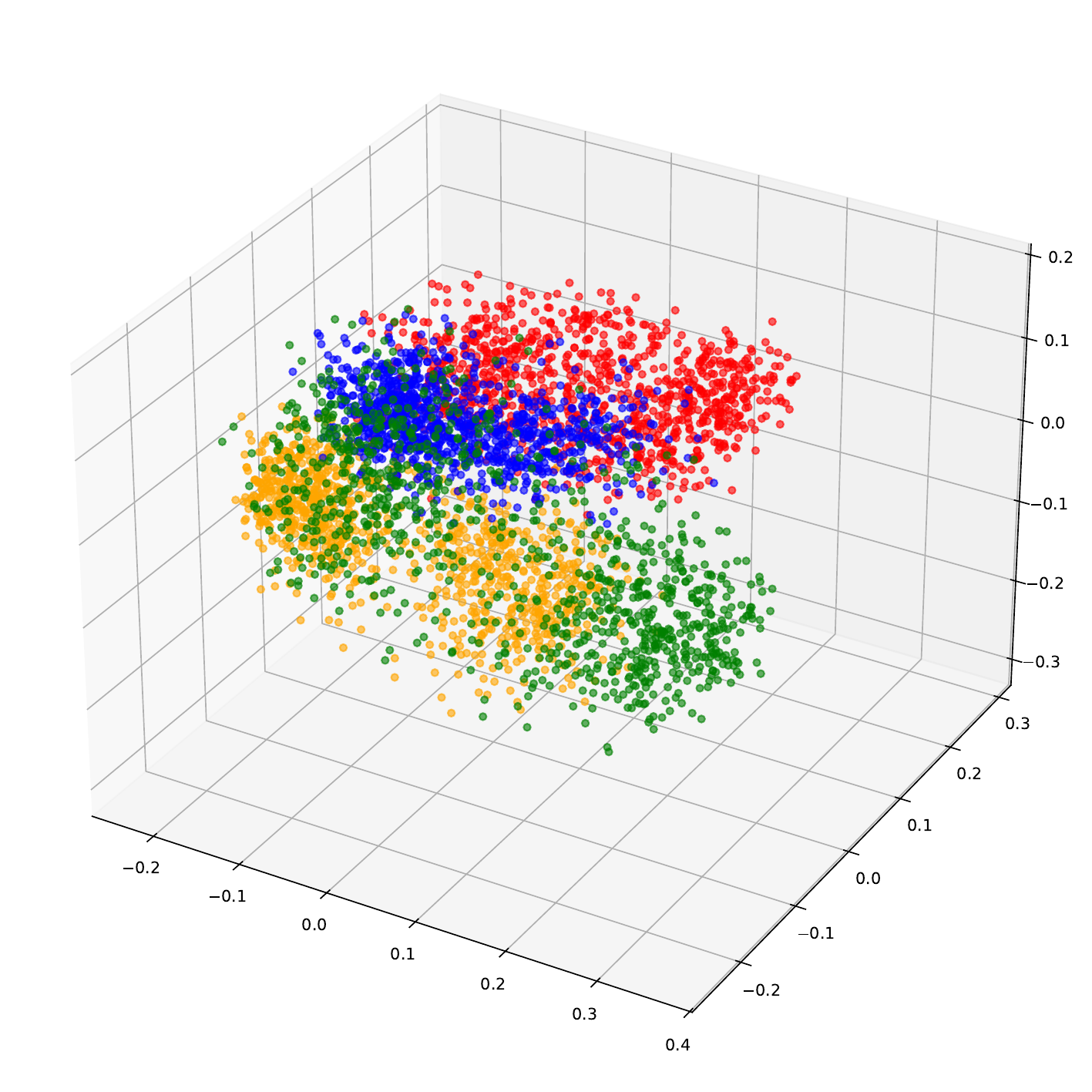}
        \caption{$m^2$-mix}
        \label{fig:PCA-neg-neg c}
    \end{subfigure}
    \begin{subfigure}{.24\linewidth}
        \centering
        \includegraphics[width=\linewidth]{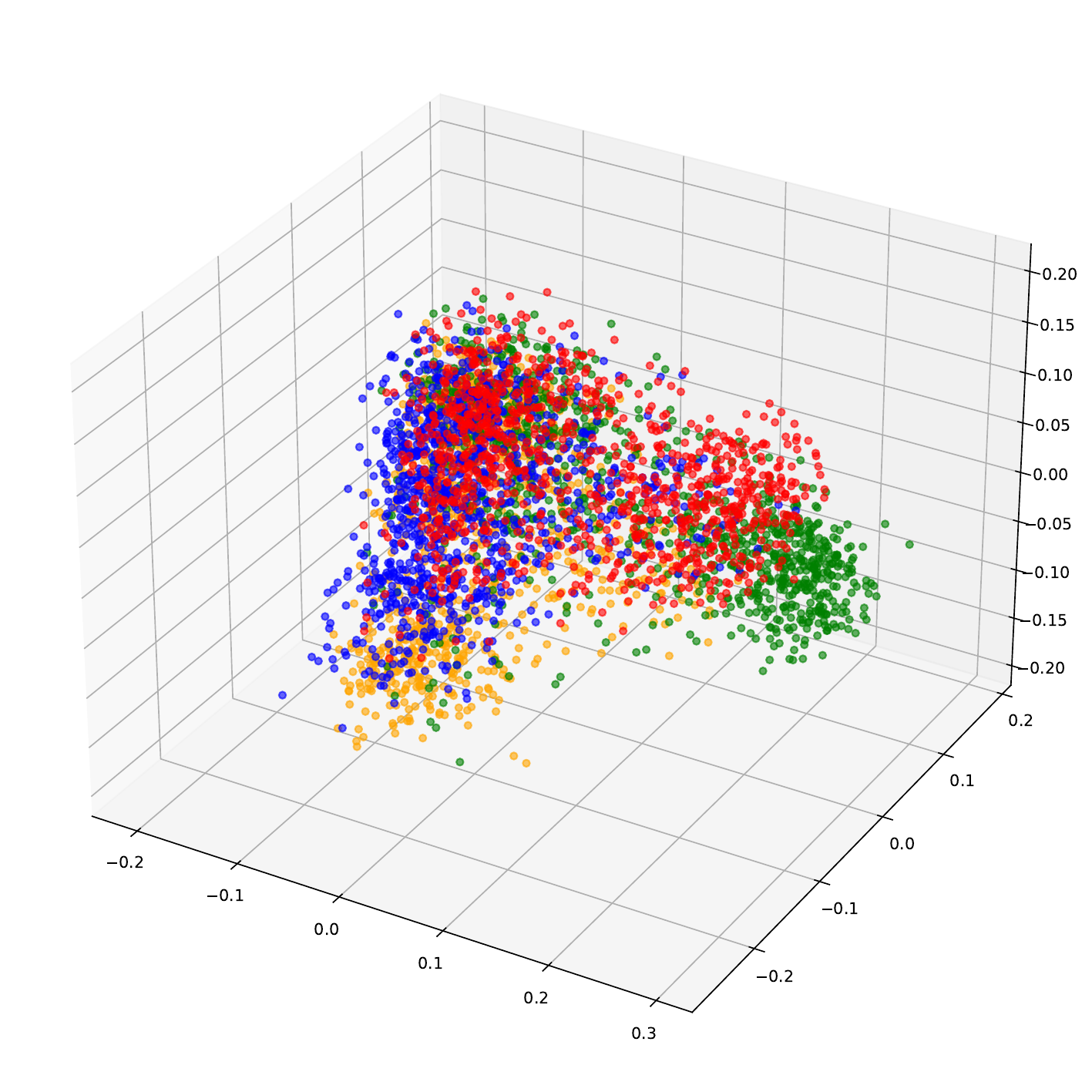}
        \caption{CMA}
        \label{fig:PCA-neg-neg d}
    \end{subfigure}
    \hspace{-2mm} 

    \caption{Visualization of image and text embeddings using PCA on ImageNet-1k, MOS benchmark datasets, and negative texts. Orange and blue points represent ID image and ID text embeddings, respectively, while green and red points denote OoD image and negative text embeddings.
    }\label{fig:PCA-neg-neg}
\end{figure*}

To achieve more intuitive visualization, we use PCA with 1,000 image prototypes and class prototypes from ImageNet-1k, along with 1,000 random OoD images (from the MOS benchmark datasets) and negative texts.
As shown in Fig.~\ref{fig:PCA}, methods such as ZS and FLYP exhibit a clear modality gap between ID image and ID text embeddings (orange and blue points). This gap is also observed in OoD image and text embeddings, as illustrated in Figs.~\ref{fig:PCA-neg} and \ref{fig:PCA-neg-neg}. 

Our findings indicate that eliminating this modality gap among ID embeddings is essential for fully leveraging textual information, such as negative concept texts, as discussed in Section 5. In CMA, which addresses this modality gap, the orange and blue points are clustered closer together, as are the red and green points.

\clearpage

\end{document}